\documentclass[journal]{IEEEtran}  

\usepackage{graphicx} 
\usepackage{amsmath}
\usepackage{amssymb}  
\usepackage{algorithm, algorithmic}
\newtheorem{theorem}{Theorem}
\newtheorem{lemma}[theorem]{Lemma}

\newtheorem{definition}[theorem]{Definition}	

\usepackage{color}
										
\usepackage[hidelinks]{hyperref} 
\usepackage{multirow}

\newcommand{\rev}[1]{\textcolor{black}{#1}}
\newcommand{\mpcomment}[1]{\textcolor{black}{#1}}

\title{Consensus Complementarity Control for Multi-Contact MPC}

\author{Alp Aydinoglu$^{1}$, Adam Wei$^{2}$, Wei-Cheng Huang$^{1}$, and Michael Posa$^{1}$
\thanks{*This work was supported by the National Science Foundation under Grant No. EFRI-1935294 and CMMI-1830218.
}
    \thanks{$^{1}$Alp Aydinoglu, Wei-Cheng Huang and Michael Posa are with the General Robotics, Automation, Sensing and Perception (GRASP) Laboratory, University of Pennsylvania, USA. $^{2}$Adam Wei is with the Department of Electrical and Computer Engineering, University of Toronto, Canada.
   {\tt\small \{alpayd, wchuang, posa\}@seas.upenn.edu, adam.wei@mail.utoronto.ca} (\textit{Corresponding author: Alp Aydinoglu})}%
}

\begin{document}
\maketitle


\begin{abstract}
We propose a hybrid model predictive control algorithm, consensus complementarity control (C3), for systems that make and break contact with their environment.
Many state-of-the-art controllers for tasks which require initiating contact with the environment, such as locomotion and manipulation, require \emph{a priori} mode schedules or are too computationally complex to run at real-time rates. We present a method based on the alternating direction method of multipliers (ADMM) that is capable of high-speed reasoning over potential contact events.
Via a consensus formulation, our approach enables parallelization of the contact scheduling problem.
We validate our results on five numerical examples, including four high-dimensional frictional contact problems, and a physical experimentation on an underactuated multi-contact system. We further demonstrate the effectiveness of our method on a physical experiment accomplishing a high-dimensional, multi-contact manipulation task with a robot arm.
\end{abstract}


\IEEEpeerreviewmaketitle

\section{Introduction}

\IEEEPARstart{F}{O}{R} many important tasks such as manipulation and locomotion, robots need to make and break contact with the environment. 
Even though such multi-contact systems are common, they are notoriously hard to control. 
The main challenge is finding policies and/or trajectories that explicitly consider the interaction of the robot with its environment in order to enable stable, robust motion.
For a wide range of problems, it is computationally challenging to discover control policies and/or trajectories \cite{posa2014direct, pang2022global, shirai2022simultaneous, onol2019contact, winkler2018gait, cheng2022contact, raghunathan2022pyrobocop} and existing methods are not capable of running at real-time rates for complex problems.

\begin{figure}[t!]
	\hspace*{0.3cm}
	\includegraphics[width=0.9\columnwidth]{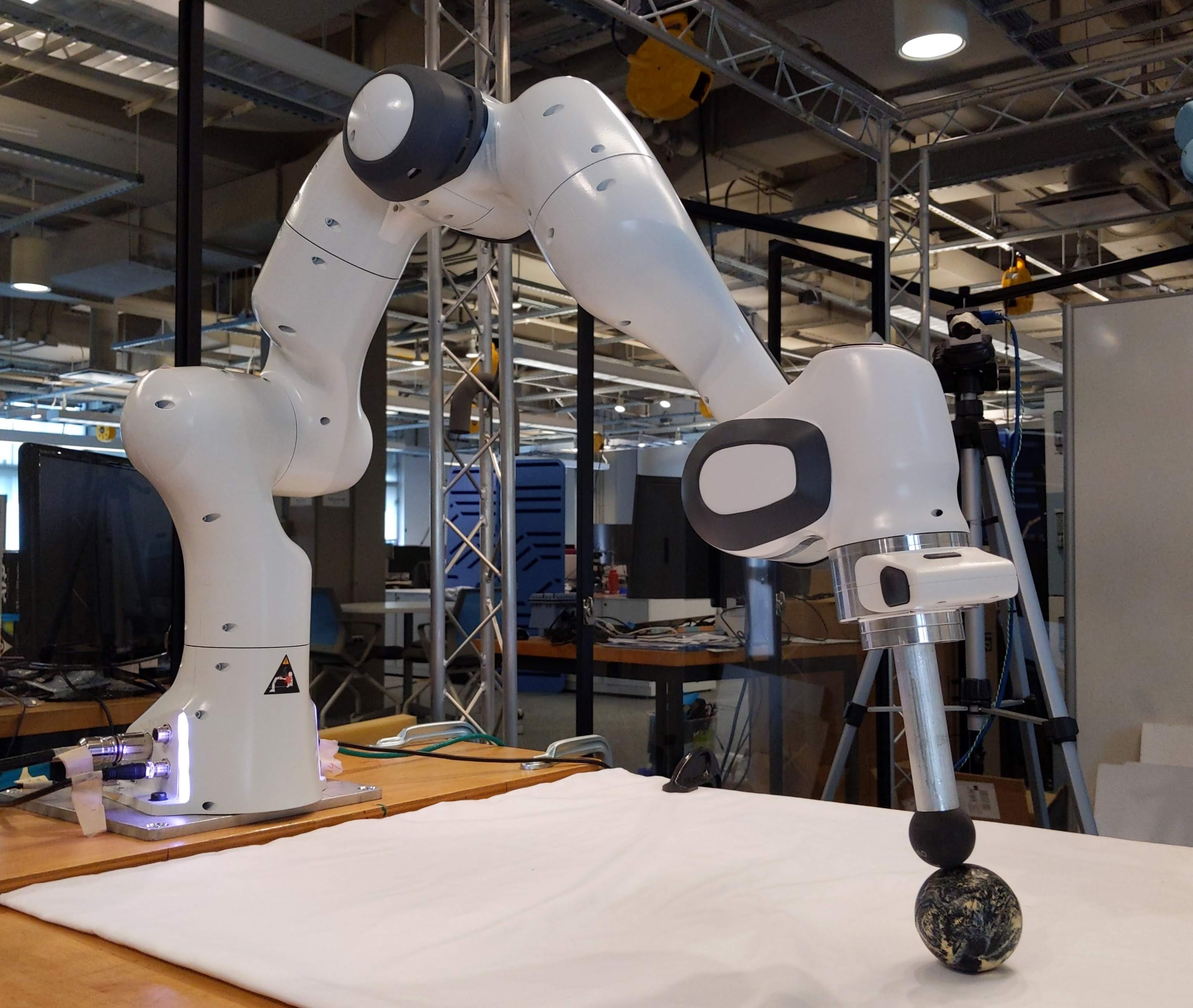}
	\caption {Manipulating a rigid object (sphere) with a spherical end-effector attached to the Franka Emika Panda arm.}
	\label{franka_ball_hw}
\end{figure}

Within robotics, model predictive control (MPC) \cite{garcia1989model, borrelli2017predictive} has become a predominant tool for automatic control, most commonly via linearization of the governing dynamics leading to quadratic programs which can be solved efficiently (e.g. \cite{ding2019real, zhakatayev2017successive} and many others). However, for multi-contact systems, the algorithm must also decide when to initiate or break contact, and when to stick or when to slide.
These discrete choices are critical to tasks like dexterous manipulation, where modeling error or disturbances can easily perturb any pre-planned or nominal contact sequence, requiring real-time decision making to adjust and adapt. 
For example, for a robot arm to roll a spherical object across a surface (Figure~\ref{franka_ball_hw}), it must alternate between rolling and repositioning the hand while reacting to adversarial disturbances; for multi-object manipulation problems, the tasks becomes exponentially more difficult, as the robot must also decide which object or objects to touch at any given instant.

The result is a \textit{hybrid} control problem. The resulting multi-modal dynamics are non-smooth and fundamentally cannot be accurately captured by linearization.
MPC for these problems has classically focused on the use of abstract representations for hybrid systems, where discrete variables encode switching between modes, and continuous variables describe motion within modes. 
The result is often encoded via mixed-integer formulations, for instance as a mixed-integer quadratic program (MIQP)  \cite{bemporad1999control}, \cite{marcucci2020warm}.
However, due to the combinatoric complexity of reasoning over potential mode sequences, these methods struggle to scale to realistic robotics problems and real-time rates.

Rather than start with a generic hybrid representation, we instead focus on the dynamics of multi-contact robotics, paired with ADMM algorithms \cite{boyd2011distributed, stellato2018embedded,takapoui2020simple, park2017general, frick2019low}. 
Specifically, we locally approximate multi-modal dynamics as a linear complementarity system (LCS) \cite{heemels2000linear}, and design an ADMM-based algorithm which can take advantage of the structure of the resulting multi-contact dynamics.

The primary contribution of this paper is an algorithm, consensus complementarity control (C3), for solving the hybrid MPC problem approximately for multi-contact systems. We exploit the distributed nature of ADMM and demonstrate that the hard part of the problem, reasoning about contact events, can be parallelized. This enables our algorithm to be fast, 
robust to disturbances and also minimizes the effect of control horizon on the run-time of the algorithm.

A preliminary version of this article was presented at the International Conference on Robotics and Automation \cite{aydinoglu2022real}. \rev{In this work, extensions are as follows.
\begin{enumerate}
	\item For the first time, we demonstrate that solving complex robotic manipulation tasks on hardware via real-time contact-implicit MPC is possible.
	\item To achieve this, we show that multi-contact manipulation problems can be formulated as a linear complementarity system that is amenable to real-time MPC (via C3).
	\item In addition, we propose a control framework that integrates C3 with a low-level impedance controller and a vision-based state estimation setup.
	\item We introduce a novel convex projection that can improve the solve times for the C3 algorithm by more than 3 times as shown in Section \ref{subsec:multi_ball}. We also include theoretical analysis of this approximate projection, connecting its limiting behavior to linear complementarity problems.
	\item The effectiveness of the method is demonstrated on two new numerical examples of high-dimensional manipulation tasks. 
\end{enumerate}
}

\section{Related Work}

Multi-contact robotics has drawn growing interest over the last decade \cite{wensing2022optimization, suomalainen2022survey}. Early methods for both planning and control were focused on known contact sequences and hierarchical planners \cite{park2007convex, abe2007multiobjective}. Over the course of the last decade, contact-implicit planners were developed (originally for offline motion synthesis \cite{posa2014direct}) and this body of work is growing rapidly \cite{onol2019contact},  \cite{manchester2020variational}. These were first paired with controllers designed to track the planned mode (and limited to reasoning about that specific mode sequence) \cite{mastalli2020crocoddyl, sleiman2021unified}. 

Other approaches achieved real-time hybrid planning via utilizing task-specific model simplifications. Some of these methods specialize in legged systems \cite{bledt2020regularized, kuindersma2016optimization, grandia2022perceptive} and focus on simplified dynamics models tailored for locomotion tasks, often times combined with gait-scheduling heuristics. However, these methods rely heavily on application-specific model simplifications and can not be applied across a wide range of robotics systems. 
Alternative methods rely on an offline phase to reduce the search space. This can lead to efficient real-time planning, but it requires a lengthy offline phase for each problem instance \cite{marcucci2017approximate, hogan2020reactive, cauligi2021coco, zhu2022efficient}. 
Others approximate the dynamics via smooth contact models and modify the dynamics (e.g. diagonal approximation of Delassus operator) to enable high performance \cite{tassa2012synthesis, koenemann2015whole, kumar2014real}. While such methods can perform well in simulation, they have not been validated on hardware performing dynamic tasks that require constantly making and breaking contact.



Building upon this growing body of literature, this paper presents an approach to real-time generic multi-contact MPC. In the preliminary version of this paper \cite{aydinoglu2022real}, and in the related work \cite{cleac2021linear}, there has been very recent progress in developing fast contact-implicit MPC algorithms. While contact-implicit trajectory optimization approaches use global models and can take minutes to solve \cite{posa2014direct, onol2019contact}, these novel approaches focus on local hybrid models to significantly reduce the computational cost of optimal control. Le Cleac'h/Howell et al. \cite{cleac2021linear} use a softened contact model in combination with a custom primal-dual interior-point solver to achieve real-time rates. They focus on tracking a pre-computed trajectory, which provides a nominal mode sequence to track (although their algorithm can adapt this sequence online). In contrast, our approach is demonstrated to be fully capable of mode sequence synthesis without the need for a nominal trajectory or any other offline computations.


\section{Background}
\label{section:background}
We first introduce some of the definitions and notation used throughout
this work. The function $\mathcal{I}_\mathcal{A}$ is the $0-\infty$ indicator function for an arbitrary set $\mathcal{A}$ such that $\mathcal{I}_\mathcal{A}(z) = 0$ if $z \in \mathcal{A}$ and $\mathcal{I}_\mathcal{A}(z) = \infty$ if $z \notin \mathcal{A}$. $\mathbb{N}_0$ denotes natural numbers with zero. For a positive semi-definite matrix $Q$, $Q^{1/2}$ denotes a matrix $P$ such that $P P = P^T P = Q$. The notation $\mathbf{blkdiag}(A_0, \ldots, A_{N})$ denotes a block diagonal matrix with entries in the given order (from top left ($A_0$) to bottom right ($A_{N}$)). The notation $\mathbf{diag}(a_0, \ldots, a_{N})$ is used if entries $a_i$ are scalars. For two vectors $a \in \mathbb{R}^m$ and $b \in \mathbb{R}^m$, $0 \leq a \perp b \geq 0$ denotes $a \geq 0$, $b \geq 0$, $a^Tb = 0$.

\subsection{Linear Complementarity Problem}

In this work, we utilize complementarity problems to represent the contact forces. These models are common in modeling multi-contact robotics problems \cite{stewart2000implicit} and can also be efficiently learned from data \cite{jin2022learning}.
The theory of linear complementarity problems (LCP) is well established \cite{cottle2009linear}.
\begin{definition} \label{def: LCS_definition}
	Given a vector $q \in \mathbb{R}^m$, and a matrix ${F \in \mathbb{R}^{m \times m}}$, the $\text{LCP}(q,F)$  describes the following mathematical program:
	\begin{alignat}{2}
		\label{LCS_definiton}
		\notag & \underset{}{\text{find}} && \lambda \in \mathbb{R}^m \\
		\notag & \text{subject to}  \quad && y = F \lambda + q,\\
		\notag & \quad && 0 \leq \lambda \perp y \geq 0.
	\end{alignat}
\end{definition}
Here, the vector $\lambda$ typically represents the contact forces and slack variables \cite{stewart2000implicit}, $y$ represents the gap function and orthogonality constraint embeds the hybrid structure.

\subsection{Linear Complementarity System}

We use linear complementarity systems (LCS) as local representations of multi-contact systems \cite{aydinoglu2020stabilization}. An LCS \cite{heemels2000linear} is a differential/difference equation coupled with a variable that is the solution of an LCP.

\begin{definition}
	A linear complementarity system describes the  trajectories $( x_k )_{k \in \mathbb{N}_{0}}$ and $( \lambda_k )_{k \in \mathbb{N}_{0}}$ for an input sequence $( u_k )_{k \in \mathbb{N}_{0}}$ starting from $x_0$ such that
	\begin{equation}
	\label{eq:LCS}
	\begin{aligned}
	& x_{k+1} = A x_k + B u_k + D\lambda_k + d,\\
	& 0 \leq \lambda_k \perp Ex_k +  F \lambda_k + H u_k + c \geq 0,
	\end{aligned}
	\end{equation}
	where $x_k \in \mathbb{R}^{n_x}$, $\lambda_k \in \mathbb{R}^{n_{\lambda}}$, $u_k \in \mathbb{R}^{n_u}$, $A \in \mathbb{R}^{n_x \times n_x}$, $B \in \mathbb{R}^{n_x \times n_u}$, $D \in \mathbb{R}^{n_x \times n_\lambda}$, $d \in \mathbb{R}^{n_x}$, $E \in \mathbb{R}^{n_{\lambda} \times n_x}$, $F \in \mathbb{R}^{n_\lambda \times n_\lambda}$, $H \in \mathbb{R}^{n_\lambda \times n_u}$ and $c \in \mathbb{R}^{n_\lambda}$. Equation \eqref{eq:LCS} is often called as generalized LCS \cite{pang2008differential} or inhomogeneous LCS \cite{camlibel2007lyapunov} due to existence of the vectors $c$ and $d$, but we use LCS for brevity.
\end{definition}

Vector $x_k$ represents the state and often consists of the generalized positions $q_k$ and velocities $v_k$.
For a given $k$, $x_k$ and $u_k$, the corresponding complementarity variable $\lambda_k$ can be found by solving $\text{LCP}(E x_k + H u_k + c, F)$ (see Definition \ref{def: LCS_definition}). Similarly, $x_{k+1}$ can be computed using the first equation in \eqref{eq:LCS} when $x_k, u_k$ and $\lambda_k$ are known.

For the frictional systems we study, $\text{LCP}$s can lack solutions or have multiple solutions. Here, we assume that the mapping $(x_k, u_k) \mapsto x_{k+1}$ is unique 
even though $\lambda_k$ is not necessarily unique \cite{aydinoglu2020stabilization}. This is a common assumption in contact-implicit trajectory optimization and is an outcome from relaxed simulation approaches (MuJoCo \cite{todorov2012mujoco}, Dojo \cite{howell2022dojo} etc.). This is a necessary assumption for MPC to not immediately become a robust control problem reasoning over every possible outcome.

\section{Multi-Contact Dynamics and Local Approximations}

\label{sec:model}

Before presenting the core algorithmic contributions of this paper, we describe how multi-contact robot dynamics can be approximated by an LCS. Rigid-body systems with contact can be modeled by the manipulator equation:
\begin{equation}
	\label{eq:manipulator_equation}
	M(q) \dot{v} + C(q,v) = Bu + J(q)^T \lambda,
\end{equation}
where $q$ is the generalized positions vector, $v$ is the generalized velocities, $\lambda$ represents the contact forces (potentially includes slack variables to capture frictional contact \cite{stewart2000implicit}), $M(q)$ is the inertia matrix, $C(q,v)$ represents the combined Coriolis, centrifugal and gravitational terms, $B$ maps control inputs to joint coordinates and $J(q)$ is the projection matrix that is typically the contact Jacobian (potentially padded with zeros for the corresponding slack variables).

One approach to represent $\lambda$ is via the complementarity framework:
\begin{equation}
	\label{eq:manipulator_complementarity_constraints}
	0 \leq \lambda \perp \psi (q,v, u, \lambda) \geq 0.
\end{equation}
Here $\psi$ relates the position $q$, velocity $v$ and input $u$ with the vector $\lambda$ that includes the contact forces (\cite{stewart2000implicit, brogliato1999nonsmooth} for more details). For compactness, we denote a nonlinear multi-contact model as $\mathcal{M}$ where $\mathcal{M} = \{ M, C, B, J, \psi\}$.

In this work, we are interested in solving the following optimal control problem with model $\mathcal{M}$ (we consider discrete-time formulations as in \cite{posa2014direct}):
\begin{align}
\label{eq:cost_nonlinear_opt}
\min_{q_k, v_k, u_k, \lambda_k} \quad & \sum_k g_c(q_k, v_k, u_k) \\
\label{eq:constraint_nonlinear_opt}
\textrm{s.t.} \quad &  (q_{k+1}, v_{k+1}, q_{k}, v_{k}, u_{k}, \lambda_{k}) \; \text{respects} \; \mathcal{M}
\end{align}
where $g_c$ is the cost function (often quadratic). \rev{Now, we introduce two different approaches to transcribe the dynamics.}

\rev{
\subsection{Stewart-Trinkle Formulation}	
In most of the examples in this paper, we transcribe \eqref{eq:constraint_nonlinear_opt} using the following implicit time-stepping scheme based on Stewart and Trinkle's seminal work \cite{stewart2000implicit} due to its intuitive, simple form:
}
\begin{equation}
\label{eq:ST_dynamics_nonlinear}
\begin{aligned}
q_{k+1} = q_k &+ \Delta t v_{k+1}, \\
v_{k+1} = v_k &+ \Delta t M^{-1}(q_k) \bigg( C(q_k,v_k) + B u_k \\
& + J_n(q_k)^T \lambda_k^n + J_t(q_k)^T \lambda_k^t \bigg),
\end{aligned}
\end{equation} 
where $\Delta t$ is the discretization time-step, $\lambda^n_k \in \mathbb{R}^{n_c}$ represents the normal forces, $\lambda^t_k \in \mathbb{R}^{ n_c n_e}$ represents the tangential (friction) forces, $n_c$ is the number of rigid body pairs that can interact, $n_e$ represents the number of edges of the polyhedral approximation of the friction cone, and $J_n$, $J_t$ are contact Jacobians for normal and tangential directions. The forces $\lambda^n_k$ and $\lambda^t_k$ can be described via the following complementarity problem \cite{stewart2000implicit}:
\rev{\begin{equation}
\label{eq:ST_constraints_nonlinear}
\begin{aligned}
& 0 \leq \gamma_k \perp \mu \lambda^n_k - E_t \lambda_k^t \geq 0, \\
& 0 \leq \lambda_k^n \perp \phi(q_k) + J_n(q_k) (q_{k+1} - q_k)  \geq 0, \\
& 0 \leq \lambda_k^t \perp E_t^T \gamma_k + J_t(q_k) v_{k+1} \geq 0,
\end{aligned}
\end{equation}}
where $\gamma_k \in \mathbb{R}^{n_c}$ is a slack variable, $\mu = \mathbf{diag}(\mu_1, \ldots, \mu_{n_c})$ represents the coefficient of friction between rigid body pairs, $ E_t = \mathbf{blkdiag}(e, \ldots, e)$ with $e =[1, \ldots,  1] \in \mathbb{R}^{1 \times n_e}$ and $\phi$ represents the distance between rigid body pairs.

The dynamics formulation in \eqref{eq:ST_dynamics_nonlinear} and \eqref{eq:ST_constraints_nonlinear}, as a nonlinear complementarity problem, has been employed for trajectory optimization of multi-contact robotics [1], though such methods have been limited to offline motion planning due to the inherent complexity in the nonlinear model.
To reduce the complexity of the model, it is possible to locally approximate $\mathcal{M}$ with an LCS. Even though an LCS approximation is slightly less complex than the nonlinear model, it still captures the multi-modal nature of the problem and enables one to make decisions such as making or breaking contact (as demonstrated in Sections \ref{sec:examples_low} and \ref{sec:examples_high}). Hence we focus on LCS approximations as they lead to more tractable optimal control problems than their nonlinear counterparts but note that these are still hybrid optimal control problems that are difficult to solve (Section \ref{sec:ADMM_main}).

Now, we present an approach to obtain an LCS approximation given the complementarity system model \eqref{eq:ST_dynamics_nonlinear}-\eqref{eq:ST_constraints_nonlinear}. 
The use of LCS approximations is not new to this work \cite{aydinoglu2020stabilization, cleac2021linear, marcucci2020warm}, though we note there is no single canonical LCS formulation. Here, we present an approach inspired by Stewart and Trinkle \cite{stewart2000implicit} in detail.
For a given state $(x^*)^T = [(q^*)^T, (v^*)^T]$ and input $u^*$, we approximate \eqref{eq:ST_dynamics_nonlinear} as:
\begin{equation}
	\label{eq:ST_dynamics}
	\begin{aligned}
	& q_{k+1} = q_k + \Delta t v_{k+1}, \\
	& v_{k+1} = v_k + \Delta t (  J_f \begin{bmatrix} q_k \\ v_k \\ u_k \end{bmatrix} + D_1 \lambda_k^n + D_2 \lambda_k^t + d_v).
	\end{aligned}
\end{equation} 
Here $J_f = J_f(q^*, v^*, u^*)$ is the Jacobian of $f(q,v,u) = M^{-1}(q) B u - M^{-1}(q) C(q,v)$ evaluated at $(q^*, v^*, u^*)$, $d_v = f(q^*, v^*, u^*) - J_f \begin{bmatrix} (q^*)^T & (v^*)^T & (u^*)^T \end{bmatrix}^T$ is a constant vector, $D_1 = M^{-1}(q^*) J_n(q^*)^T$ and $D_2 = M(q^*)^{-1} J_t(q^*)^T$ represent the effect of contact forces. Similarly, we approximate \eqref{eq:ST_constraints_nonlinear} with the following LCP:
\rev{
\begin{equation}
	\label{eq:ST_constraints}
	\begin{aligned}
	& 0 \leq \gamma_k \perp \mu \lambda^n_k - E_t \lambda_k^t \geq 0, \\
	& 0 \leq \lambda_k^n \perp \phi(q^*) + J_n(q^*) q_k + \Delta t J_n(q^*) v_{k+1} \\ 
	& \qquad \qquad \qquad \qquad \qquad \qquad - J_n(q^*) q^* \geq 0, \\
	& 0 \leq \lambda_k^t \perp E_t^T \gamma_k + J_t(q^*) v_{k+1} \geq 0.
	\end{aligned}
\end{equation}
}
Observe that \eqref{eq:ST_dynamics} and \eqref{eq:ST_constraints} can be written in the LCS form \eqref{eq:LCS} where $x_k^T = [q_k^T, v_k^T]$ and $\lambda_k^T = [\gamma_k^T, (\lambda_k^n)^T, (\lambda_k^t)^T]$. 
\rev{\subsection{Anitescu Formulation}
\label{subsec:Anitescu}
In Section \ref{sec:C3}, we propose a novel, approximate projection method that requires convexity of the contact model (i.e. $F = F(x_k)$ as in \eqref{eq:LCS} is positive semi-definite for any $x_k$) but the Stewart-Trinkle formulation does not satisfy this assumption. Hence, we alternatively employ a formulation from Anitescu \cite{anitescu2006optimization}, which does guarantee that $F \succeq 0$:
\begin{equation}
		\label{eq:Anitescu_dynamics_nonlinear}
		\begin{aligned}
			q_{k+1} = q_k &+ \Delta t v_{k+1}, \\
			v_{k+1} = v_k &+ M^{-1}(q_k) \bigg( \Delta t B u_k - \Delta t C(q_k,v_k)  \\
			&  \qquad \qquad \qquad \qquad + J_c(q_k)^T \lambda_k \bigg),
		\end{aligned}
\end{equation}
where the contact Jacobian is defined as $J_c(q_k) = E_t^T J_n(q_k) + \mu J_t(q_k)$. Via the complementarity equations, $\lambda_k$ is described as:
\begin{equation}
		\label{eq:Anitescu_constraints_nonlinear}
				\begin{aligned}
		0 \leq \lambda_k \perp \frac{E_t^T \phi(q_k)}{\Delta t} + \frac{1}{\Delta t} E_t^T &J_n (q_k) (q_{k+1} - q_k) \\ 
		&+ \mu J_t(q_k) v_{k+1} \geq 0.
		\end{aligned}
\end{equation}
Similarly (for the Anitescu formulation), given state $(x^*)^T = [(q^*)^T, (v^*)^T]$ and input $u^*$, we can approximate \eqref{eq:Anitescu_dynamics_nonlinear} as:
\begin{equation}
	\label{eq:Anitescu_linear_dynamics}
	\begin{aligned}
		& q_{k+1} = q_k + \Delta t v_{k+1}, \\
		& v_{k+1} = v_k + \Delta t J_f \begin{bmatrix} q_k \\ v_k \\ u_k \end{bmatrix} + D \lambda_k + \Delta t d_v.
	\end{aligned}
\end{equation}
where $J_f$, $d_v$ are same as in \eqref{eq:ST_dynamics} and $D = M^{-1}(q^*) J_c(q^*)^T$. The complementarity part \eqref{eq:Anitescu_constraints_nonlinear} is:
\begin{equation}
	\label{eq:Anitescu_linear_constraints}
	\begin{aligned}
	0 \leq \lambda_k \perp \frac{1}{\Delta t}E_t^T \bigg( \phi(q^*) + J_n(q^*) &q_k - J_n(q^*) q^* \bigg) \\
	& + J_c(q^*) v_{k+1}  \geq 0.
	\end{aligned}
\end{equation}
Similar to the LCS approximation of the Stewart-Trinkle formulation, \eqref{eq:Anitescu_linear_dynamics} and \eqref{eq:Anitescu_linear_constraints} can be written in the LCS form \eqref{eq:LCS} where $x_k^T = [q_k^T, v_k^T]$. 
While a full comparison of these methods is outside the scope of this paper, we note that convexity in the Anitescu formulation is not without cost: this version does introduce some modeling artifacts, particularly during high-speed sliding motion.
}

\rev{
It is important to note that to obtain these LCS models, we approximate the differentiable terms (of the model $\mathcal{M}$), which arise form standard Lagrangian dynamics, by linearizing with respect to $(q,v,u)$ about $(q^*, v^*, u^*)$ but leave the multi-modal structure intact.
}
Moving forward, we focus on LCS models of the form \eqref{eq:LCS} as it is a general form. We do not include time-varying LCS (where matrices such as $A$ in \eqref{eq:LCS} depend on $k$) for ease of notation but note that this work can easily be extended to the time-varying setting (and code we provide can deal with time-varying LCS). With a slight abuse of notation, we denote the models as $\mathcal{L}_{\Delta t} (x, u)$ where dependence on a given state $x^*$ and input $u^*$ (e.g. as in \eqref{eq:ST_dynamics}, \eqref{eq:ST_constraints}) is suppressed. 
Furthermore we use the notation $(x_{k+1}, \lambda_k) = \mathcal{L}_{\Delta t} (x_k, u_k)$ to denote the state of the LCS system after one time step ($\Delta t$ time later). Here, $\lambda_k$ is the solution of $\text{LCP}(E x_k + H u_k + c, F)$ and $x_{k+1} = Ax_k + Bu_k + D \lambda_k + d$.

\section{Model Predictive Control of Multi-Contact Systems}
\label{sec:ADMM_main}

As we consider LCS models, we focus on the following mathematical optimization problem:
\begin{equation}
\label{eq:MPC_original}
\begin{aligned}
\min_{x_k, \lambda_k, u_k} \quad & \sum_{k=0}^{N-1} (x_k^T Q_k x_k + u_k^T R_k u_k)  + x_N^T Q_N x_N \\
\textrm{s.t.} \quad &x_{k+1} = A x_k + B u_k + D \lambda_k + d, \\
& E x_k + F \lambda_k + H u_k + c \geq 0, \\
& \lambda_k \geq 0, \\
& \lambda_k^T (E x_k + F \lambda_k + H u_k + c) = 0, \\
& (\boldsymbol x, \boldsymbol \lambda, \boldsymbol u) \in \mathcal{C}, \\
& \text{for} \; k = 0, \ldots, N-1, \text{given} \; x_0,
\end{aligned}
\end{equation}
where $N$ is the planning horizon, $Q_k, Q_N$ are positive semidefinite matrices, $R_k$ are positive definite matrices and $\mathcal{C}$ is a convex set (e.g. input bounds, safety constraints, or goal conditions) and $\boldsymbol{x}^T = [x_1^T, \ldots, x_{N}^T]$, $\boldsymbol{\lambda}^T = [\lambda_0^T, \lambda_1^T, \ldots, \lambda_{N-1}^T]$, $\boldsymbol{u}^T = [u_0^T, u_1^T, \ldots, u_{N-1}^T]$. 

Given $x_0$, one solves the optimization \eqref{eq:MPC_original} and applies $u_0$ to the plant and repeats the process in every time step in a receding horizon manner.

\subsection{Mixed Integer Formulation}

One straightforward, but computationally expensive, approach to solving \eqref{eq:MPC_original} is via a mixed integer formulation which exchanges the non-convex orthogonality constraints for binary variables:
\begin{equation}
\label{eq:MPC_MIQP}
\begin{aligned}
\min_{x_k, \lambda_k, u_k, s_k} \quad & \sum_{k=0}^{N-1} (x_k^T Q_k x_k + u_k^T R_k u_k)  + x_N^T Q_N x_N \\
\textrm{s.t.} \quad &x_{k+1} = A x_k + B u_k + D \lambda_k + d, \\
& M s_k \geq E x_k + F \lambda_k + H u_k + c \geq 0, \\
& M (\mathbf{1} - s_k) \geq \lambda_k \geq 0, \\
& (\boldsymbol x, \boldsymbol \lambda, \boldsymbol u) \in \mathcal{C}, s_k \in \{ 0,1 \}^{n_\lambda}, \\
& \text{for} \; k = 0, \ldots, N-1, \text{given} \; x_0,
\end{aligned}
\end{equation}
where $\mathbf{1}$ is a vector of ones, $M$ is a scalar used for the big M method and $s_k$ are the binary variables. 
This approach requires solving $2^{N n_\lambda}$ quadratic programs as a worst case.
This method is popular because, in practice, it is much faster than this worst case analysis. Even still, for the multi-contact problems explored in this work, directly solving \eqref{eq:MPC_MIQP} via state-of-the-art commercial solvers remains too slow for real-time control. Methods that learn the MIQP problem offline are promising, but this line of work requires large-scale training on every problem instance \cite{cauligi2020learning}, \cite{aydinoglu2021stability}.

\subsection{Consensus Complementarity Control (C3)}
\label{sec:C3}

Utilizing a method based on ADMM, we will solve \eqref{eq:MPC_original} more quickly than with mixed integer formulations.
Since the problem we are addressing is non-convex, our method is not guaranteed to find the global solution or converge unlike MIQP-based approaches, but is significantly faster.

\rev{Towards this direction, we first transform problem \eqref{eq:MPC_MIQP} into a consensus formulation. Then, we apply ADMM to solve the problem written in consensus form \eqref{eq:MPC_consensus} by transforming it into a set of three distinct operations. Finally, we propose four different approaches to solve one of those operations (the projection operation). }

\subsubsection{Consensus Formulation}
First, we rewrite \eqref{eq:MPC_original}, equivalently, in the consensus form \cite{park2017general} where we create copies (named $\delta_k$) of variables $z_k^T = [x_k^T, \lambda_k^T, u_k^T]$ and move the constraints into the objective function using $0 -\infty$ indicator functions:
\begin{equation}
	\label{eq:MPC_consensus}
	\begin{aligned}
		\min_{ z } \quad & c(z) + \mathcal{I}_\mathcal{D} ( z  ) + \mathcal{I}_\mathcal{C} ( z  ) + \sum_{k=0}^{N-1} \mathcal{I}_{\mathcal{H}_k} (\delta_k) \\
		\textrm{s.t.} \quad &z_k = \delta_k, \; \forall k,
	\end{aligned}
\end{equation}
where $z^T = [z_0^T, z_1^T, \ldots, z 	_{N-1}^T]$. Note that $\delta^T = [\delta_0^T, \delta_1^T, \ldots, \delta_{N-1}^T]$ is a copy of $z^T$ and equality constraints can also be written as $z = \delta$. $c(z)$ is the cost function\footnote{The cost function has the form $c(z) =  \sum_{k=0}^{N-1}(x_k^T Q_k x_k + u_k^T R_k u_k) + || Q_N^{1/2} (Ax_{N-1} + B u_{N-1} + D \lambda_{N-1} + d) ||_2^2$.
	
} in \eqref{eq:MPC_original}.
The set $\mathcal{D}$ includes the dynamics constraints:
\begin{equation*}
	\cap_{k=0}^{N-2} \{ z : x_{k+1} = A x_k + B u_k + D \lambda_k + d  \},
\end{equation*}
and the sets $\mathcal{H}_k$ represent the LCP (contact) constraints:
\begin{align*}
	\mathcal{H}_k = \{&(x_k, \lambda_k, u_k) : E x_k + F \lambda_k + H u_k + c \geq 0, \\
	& \lambda_k \geq 0, \lambda_k^T (E x_k + F \lambda_k + H u_k + c) = 0  \}.
\end{align*}
Note that we leverage the time-dependent structure in the complementarity constraints to separate $\delta_k$'s so that each set $\mathcal{H}_k$ depends only on $\delta_k$, and not variables corresponding to any other timestep.

\rev{\subsubsection{ADMM Steps for the Consensus Formulation}
In this part, we discuss how to solve the problem in consensus formulation \eqref{eq:MPC_consensus} via ADMM.	
}
The general augmented Lagrangian (\cite{boyd2011distributed}, Section 3.4.2.) for the problem in consensus form \eqref{eq:MPC_consensus} is (Appendix for details):
\begin{align}
	\label{eq:lagrangian}
	\notag
	\mathcal{L}_\rho & (z,  \delta, w) = c(z) + \mathcal{I}_\mathcal{D} ( z ) + \mathcal{I}_\mathcal{C} ( z )   \\
	&+ \sum_{k=0}^{N-1} \big( \mathcal{I}_{\mathcal{H}_k} (\delta_k) + \rho ( r_k^T G_k r_k - w_k^T G_k w_k) \big ),
\end{align}
where $\rho >0$ is the penalty parameter, $w^T = [w_0^T, w_1^T, \ldots, w_{N-1}^T]$, $w_k$ are scaled dual variables, $r_k = z_k - \delta_k + w_k$, $G_k$ is a positive definite matrix. Observe that the standard augmented Lagrangian \cite{park2017general} is recovered for $G_k = I$ for all $k$.

In order to solve \eqref{eq:MPC_consensus}, we apply the ADMM algorithm, consisting of the following operations (Appendix for details):
\begin{align}
	\label{eq:bir}
	& z^{i+1} = \text{argmin}_z \mathcal{L}_\rho (z, \delta^i, w^i), \\
	\label{eq:iki}
	& \delta_k^{i+1} = \text{argmin}_{\delta_k}  \mathcal{L}_\rho^k (z_k^{i+1}, \delta_k, w_k^i ), \; \forall k, \\
	\label{eq:uc}
	& w_k^{i+1} = w_k^i + z^{i+1}_k - \delta_k^{i+1}, \; \forall k,
\end{align}
where $\mathcal{L}_\rho^k(z_k, \delta_k, w_k) =  \mathcal{I}_{\mathcal{H}_k} (\delta_k) + \rho (r_k^T G_k r_k - w_k^T G_k w_k)$.
Here, \eqref{eq:bir} requires solving a quadratic program, \eqref{eq:iki} is a projection onto the LCP constraints and \eqref{eq:uc} is a dual variable update. Next, we analyze these operations in the given order.

\paragraph{Quadratic Step}
Equation \eqref{eq:bir} can be represented by the convex quadratic program
\begin{equation}
	\label{eq:quadratic_program}
	\begin{aligned}
		\min_{z} \quad & c(z) + \sum_{k=0}^{N-1} (z_k - \delta_k^i + w_k^i)^T \rho G_k (z_k - \delta_k^i + w_k^i)\\
		\textrm{s.t.} \quad & z \in \mathcal{D} \cap \mathcal{C}.
	\end{aligned}
\end{equation}
The linear dynamics constraints are captured by the set $\mathcal{D}$, and the convex inequality constraints on states, inputs, contact forces are captured by $\mathcal{C}$. 
The complementarity constraints do not explicitly appear, but their influence is found, iteratively, through the variables $\delta_k^i$.

The QP in \eqref{eq:quadratic_program} can be solved quickly via off-the-shelf solvers and is analogous to solving the MPC problem for a linear system without contact.

\paragraph{Projection Step}

This step requires projecting onto the LCP constraints $\mathcal{H}_k$ and is the most challenging part of the problem. \eqref{eq:iki} can be represented with the following quadratic program with a non-convex constraint:
\begin{equation}
	\label{eq:projection_program}
	\begin{aligned}
		\min_{\delta_k} \quad & (\delta_k - (z_k^{i+1} + w^i_k) )^T \rho G_k (\delta_k - (z_k^{i+1} + w^i_k) )  \\
		\textrm{s.t.} \quad & \delta_k \in \mathcal{H}_k,
	\end{aligned}
\end{equation}
where $\delta_k^T = [(\delta_k^x)^T, (\delta_k^\lambda)^T, (\delta_k^u)^T]$.
\rev{
Within this framework, we present four alternative algorithms for this projection stage. Three are approximate projections, common for minimization problems over non-convex sets \cite{diamond2018general}.}

\rev{
\subsubsection{Projection Operation}
\label{subsec:projection}
As stated, we propose four different methods for the projection step. The first three apply to any LCS model, while the fourth requires $F \succeq 0$, and thus is limited to the Anitescu formulation of dynamics.
}
\paragraph{MIQP Projection}
The projection can be calculated by exactly formulating \eqref{eq:projection_program} as a small-scale MIQP
\begin{equation}
	\label{eq:approx_proj}
	\begin{aligned}
		\min_{\delta_k, s_k} \quad & (\delta_k - (z_k^{i+1} + w^i_k) )^T U (\delta_k - (z_k^{i+1} + w^i_k) )  \\
		\textrm{s.t.} \quad & M s_k \geq E \delta_k^x + F \delta_k^\lambda + H \delta_k^u + c \geq 0, \\
		& M (\mathbf{1} - s_k) \geq \delta_k^\lambda \geq 0, \\
		& s_k \in \{ 0,1 \}^{n_\lambda},
	\end{aligned}
\end{equation}
where $U$ is a positive semi-definite matrix. 
For $U = \rho G_k$, one recovers the problem in \eqref{eq:projection_program};
however, in our experience, we found significantly improved performance using alternate, but fixed, choices for $U$.
Observe that while \eqref{eq:approx_proj} is non-convex, it is written only in terms of variables corresponding to a single time step $k$. While the original MIQP formulation in \eqref{eq:MPC_MIQP} has $Nn_\lambda$ binary variables, here we have $N$ independent problems, each with $n_\lambda$ binary variables. This decoupling leads to dramatically improved performance (worst-case $N 2^{n_\lambda}$ vs $2^{Nn_\lambda}$).

\paragraph{LCP Projection}
In cases where \eqref{eq:approx_proj} cannot be solved quickly enough, we propose \rev{three} approximate solutions with faster run-time.
Consider the limiting case where $U$ has no penalty on the force elements. Here, \eqref{eq:approx_proj} can be solved with optimal objective value of $0$ by setting $\delta_k^x = z_k^{(i+1),x} + w_k^{(i),x}$ and $\delta_k^u = z_k^{(i+1),u} + w_k^{(i),u}$. Then, $\delta_k^\lambda$ can be found by solving $\text{LCP}(E \delta_k^x + H \delta_k^u + c,F)$.
We note that this projection is different than shooting based methods since we are simulating the $z_k^{(i+1),x,u} + w_k^{(i),x,u}$ instead of $z_k^{(i+1),x,u}$.


\paragraph{ADMM Projection}
We note that \eqref{eq:approx_proj} has an equivalent quadratically constrained quadratic program (QCQP) representation, where prior work has solved problems of this form using ADMM (\cite{park2017general}, Section 4.4). This approach relies on the fact that QCQPs with a single constraint are solvable in polynomial time via the bisection method (\cite{park2017general}, Appendix B). Performing the projection step via this method leads to nested ADMM algorithms, but this formulation can  produce faster solutions than MIQP solvers without guarantees that it produces a feasible or optimal value. 
Note that this formulation fared poorly for high dimensional projections or when applied directly to the original problem \eqref{eq:MPC_original}, rarely satisfying the complementarity constraints.
\rev{
\paragraph{Convex Projection}
Here, we introduce a novel approximate projection that is computationally efficient, as it only requires solving a single quadratic program:
\begin{equation}
\label{eq:convex_proj}
\begin{aligned}
	\min_{\delta_k} \quad & \frac{1} {2} ||\delta_k^x - x_d||_{Q_x}  + \frac{1}{2} ||\delta_k^u - u_d||_{Q_u}  \\
	& \qquad + \frac{\alpha}{2} ||\delta_k^\lambda - \lambda_d||_{F} + \frac{1-\alpha}{2} ||\delta_k^\lambda||_{F}   \\
	\textrm{s.t.} \quad &  E \delta_k^x + F \delta_k^\lambda + H \delta_k^u + c \geq 0, \\
	&  \delta_k^\lambda \geq 0,
\end{aligned}
\end{equation}
where $x_d = z_k^{(i+1),x} + w_k^{(i),x}$, $u_d = z_k^{(i+1),u} + w_k^{(i),u}$, $\lambda_d = z_k^{(i+1),\lambda} + w_k^{(i),\lambda}$ are the desired $x$, $u$, $\lambda$ targets for the projection step respectively, and $\alpha\in[0,1]$ is a hyperparameter. The cost term in \eqref{eq:convex_proj} mimics that of the true projection, but blends in terms which approximate complementarity error.
We note the role of $F$ (from \eqref{eq:LCS}), to which we add a regularizing term $\epsilon I$ to transform the positive semi-definite $F$ from the Anitescu model into a strictly positive definite matrix.}

\rev{One benefit of this approximate projection is that, in the limit $\alpha \to 0$, any error in the complementarity constraints also approaches zero, and so $\alpha$ can be seen as a hyperparameter which regulates the complementarity violation.
}

\rev{
\begin{lemma}
\label{lemma:convex_proj_err}
	The complementarity error is bounded linearly with $\alpha$, i.e. $(\delta_k^\lambda)^T (E \delta_k^x + F \delta_k^\lambda + H \delta_k^u + c) \rightarrow 0$ as $\alpha \rightarrow 0$.
\end{lemma}
\begin{IEEEproof}
In Appendix.
\end{IEEEproof}
}

\rev{Empirically, solving the full MIQP \eqref{eq:approx_proj} typically is the best at exploring a wide range of modes, while the LCP and convex projection are usually the fastest (depending on matrix $F$). The ADMM projection can be viewed as a middle ground. We underline that both MIQP and LCP projections produce feasible solutions whereas ADMM and the convex projection have no such guarantee. For frictional contact problems (Sections \ref{sec:examples_low} and \ref{sec:examples_high}), we observed that the MIQP and convex projections performed better than the others.}



\begin{algorithm}[t!]
	\caption{Consensus Complementarity Control (C3)}
	\begin{algorithmic}[1]
		\REQUIRE $\theta = \{ Q_k, R_k, Q_N, \delta^0_k, w^0_k, G_k, s, \rho, \rho_s, N \} $, $\mathcal{L}_{\Delta t}$, $x_0$
		\\ \textit{Initialization} : $i=1$
		\WHILE {$i \leq s $}
		\STATE Compute $z^{i+1}$ via \eqref{eq:quadratic_program}
		\STATE Compute $\delta_k^{i+1}$ via \eqref{eq:approx_proj}, $\forall k$
		\STATE  $w_k^{i+1} \gets w_k^i + z^{i+1}_k - \delta_k^{i+1}, \forall k$
		\STATE $\rho \gets \rho_s \rho$
		\STATE $w_k^{i+1} \gets w_k^{i+1} / \rho_s, \forall k $
		\STATE $i \leftarrow i +1$
		\ENDWHILE
		\RETURN $u_0$ from $z^{s + 1}$
	\end{algorithmic} 
	\label{algortihm_ADMM}
\end{algorithm}

After discussing each of the individual steps, we present the full C3 algorithm (Algorithm \ref{algortihm_ADMM}). Here $\theta$ represents the C3 parameters. Both $\delta_k^0$ and $w_k^0$ are usually initialized as zero vectors. $G_k$ are positive definite matrices, $s > 0 $ is the number of ADMM steps, $\rho_s > 0$ is the scaling parameter for $\rho$, and $N$ is the number of planning steps. Furthermore $\mathcal{L}_{\Delta t}$ is the LCS model (Section \ref{sec:model}) and $x_0$ is an initial state. Given this information, the algorithm returns $u_0$ as in standard model-predictive control frameworks.

\section{Illustrative Examples}

\label{sec:examples_low}


In this section, we demonstrate the effectiveness of C3 presenting multiple simulation results and an hardware experiment with an under-actuated multi-contact system. 
Concrete steps of method used throughout this section is shown in Algorithm \ref{algortihm_low_level}. Given a multi-contact model $\mathcal{M}$, we obtain an LCS model $\mathcal{L}_{\Delta{t}}$ around the estimated state $\hat{x}$ and nominal input $\hat{u}$. Then, we run Algorithm \ref{algortihm_ADMM} to compute $u_0$ and that is directly applied to the system.

For these results, OSQP \cite{osqp} is used to solve quadratic programs and Gurobi \cite{Gurobi} is used for mixed integer programs. PATH \cite{dirkse1995path} and Lemke's algorihm have been used to solve LCPs. SI units (meter, kilogram, second) are used. The experiments are done on a desktop computer with the processor Intel \emph{i7-11800H} and \emph{16GB RAM}. 
Reported run-times include all steps in the algorithm.
The code for all examples is available\footnote{\url{https://github.com/AlpAydinoglu/coptimal}}, \rev{ and we optimized the code implementation leading to algorithm speeding up especially for the finger gaiting and pivoting examples compared to the preliminary conference version. We have added the specific systems matrices for all the examples\footnote{ \url{https://github.com/AlpAydinoglu/coptimal/blob/main/system_matrices.txt} }. }Experiments are also shown in the supplementary video\footnote{ \url{https://youtu.be/L57Jz3dPwO8} }.

\begin{figure}[b!]
	\hspace*{2cm}
	\includegraphics[width=0.6\columnwidth]{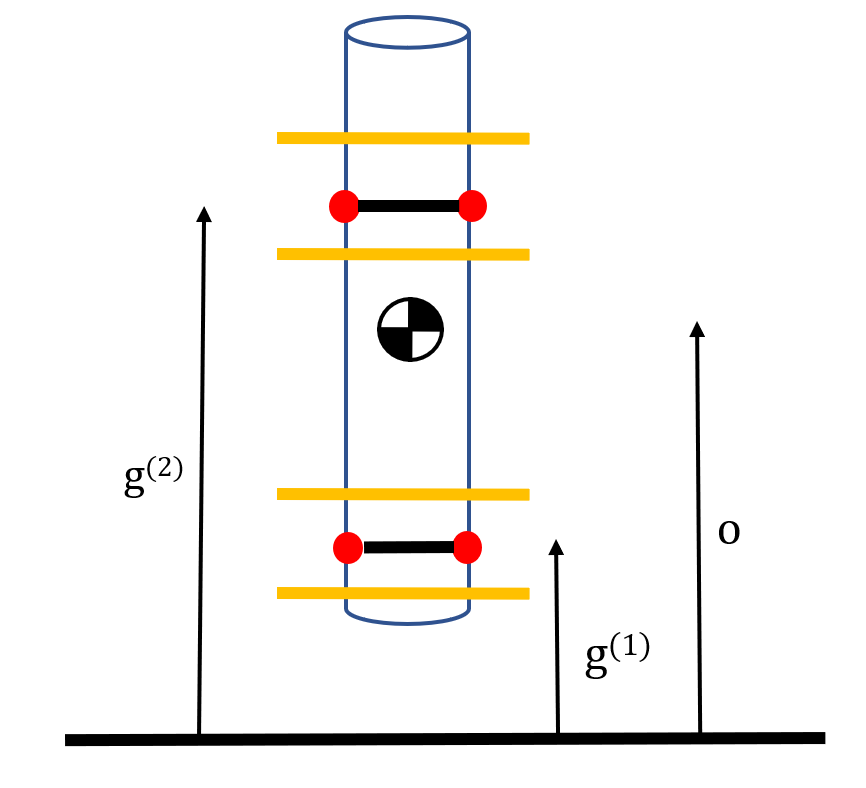}
	\caption{Lifting an object using two grippers indicated by red circles. The limits where the grippers should not cross are indicated by yellow lines.}
	\label{finger_picture}
\end{figure}

\begin{algorithm}[t!]
	\caption{}
	\begin{algorithmic}[1]
		\REQUIRE $\mathcal{M}, \theta$, $\Delta t$, $\hat{u}$
		\STATE Estimate state as $\hat{x}$
		\STATE Obtain LCS model $\mathcal{L}_{\Delta t}$ around $(\hat{x}, \hat{u})$ using $\mathcal{M}$
		\STATE Run Algorithm \ref{algortihm_ADMM} with $\mathcal{L}_{\Delta t}$, $\theta$, $\hat{x}$ and obtain $u_0$
	\end{algorithmic} 
	\label{algortihm_low_level}
\end{algorithm}

\subsection{Finger Gaiting}

In this subsection, we want to show that our algorithm is capable of mode exploration. Our goal is to lift a rigid object upwards using four fingers. The setup for this problem is illustrated in Figure \ref{finger_picture}. The red circles indicate where the grippers interact the object and we assume that the grippers are always near the surface of the object and the force they apply on the object can be controlled. This force affects the friction between the object and grippers. Since the grippers never leave the surface, we assume that there is no rotation. 
The goal of this task is to lift the object vertically, while the fingers are constrained to stay close to their original locations (constraints shown in yellow). This task, therefore, requires finger gaiting to achieve large vertical motion of the object.

We use the formulation in \cite{stewart2000implicit} for modeling the system and denote the positions of the grippers as $g^{(1)}$, $g^{(2)}$ respectively and position of the object as $o$. We choose $g=9.81$ as gravitational acceleration and $\mu=1$ is the coefficient of friction for both grippers. \rev{The system has six states ($n_x = 6$), including the position and velocity of the object and two fingers. Also, there are six complementarity variables ($n_\lambda = 6$) where each contact is represented by 3 complementarity variables. There are four inputs ($n_u = 4$), including the normal forces for finger contacts and the acceleration of the fingers.}

We design a controller based on Algorithm \ref{algortihm_ADMM} where $G_k = I$, $N = 10$, \rev{$\Delta t = 0.1$ for $\mathcal{L}_{\Delta t}$}. \rev{The controller uses the MIQP projection method since the other two projection methods failed to move the object close to the desired target.}
For this example, we use $s = 10$ and $\rho_s = 1.2$  \rev{and C3 ran at $30.3$ Hz, an  approximately $2.5$x speedup from $10$ Hz \cite{aydinoglu2022real}}. We also enforce limits:
\begin{align*}
& 1 \leq g^{(1)} \leq 3, \; \forall k, \\
& 3 \leq g^{(2)} \leq 5, \; \forall k.	
\end{align*}

\begin{figure}[t!]
	\hspace*{-0.3cm}
	\includegraphics[width=1\columnwidth]{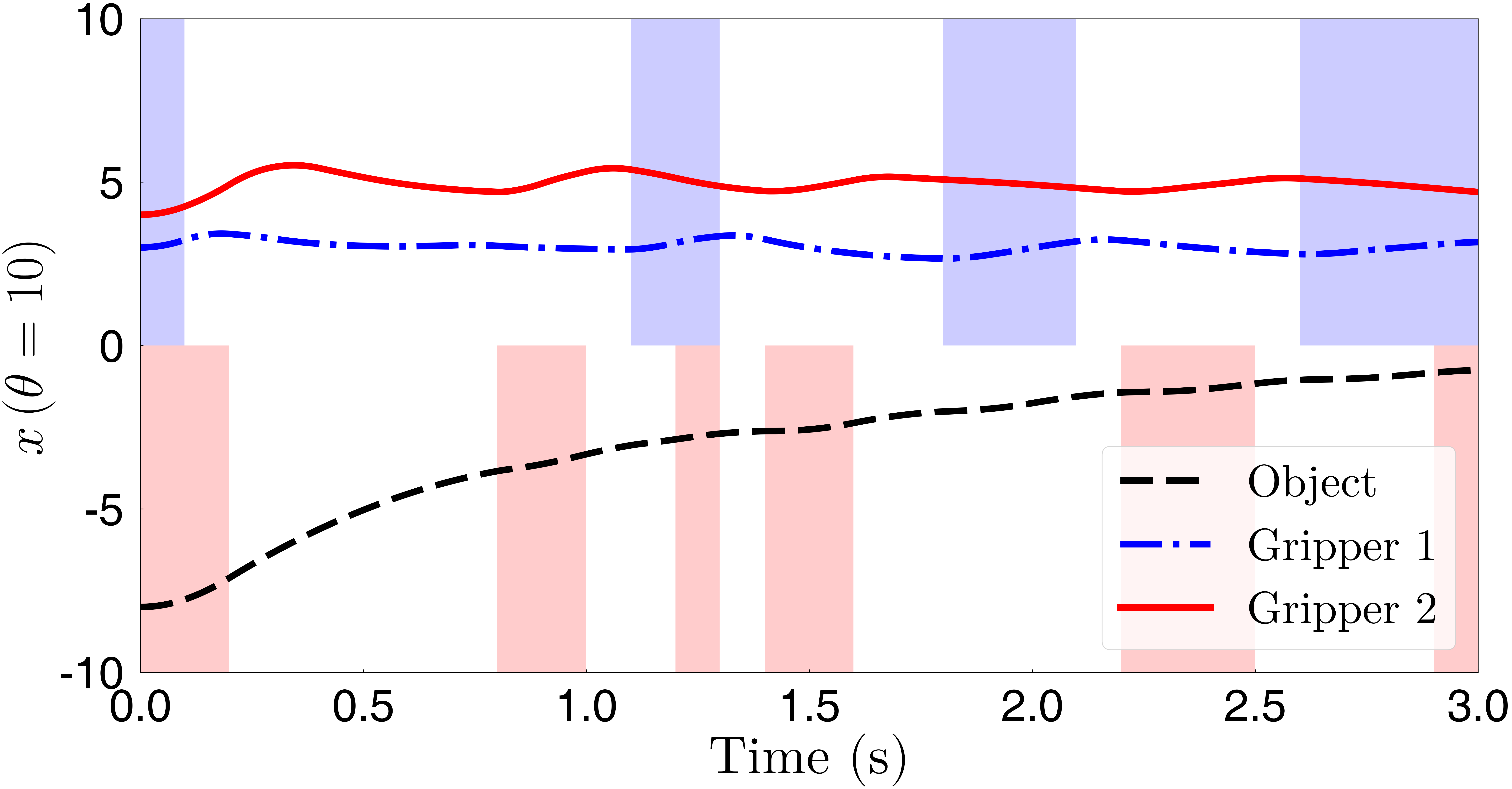}
	\caption{\rev{Finger gaiting with $s = 10$. The upper blue shading implies that gripper 1 is applying normal force to the object whereas the lower red shading implies that gripper 2 is applying normal force.}}
	\label{finger_1}
\end{figure}

We performed 100 trials starting from different initial conditions where $o(0)  \sim U[-6, -8]$, $g^{(1)}(0) \sim U[2, 3]$ , $g^{(2)}(0) \sim U[3, 4]$ are uniformly distributed and both the grippers and the object have zero initial speed. The controller managed to lift the object in all cases. We present a specific example in Figure \ref{finger_1} where the grippers first throw the object into the air and then catch it followed by some finger gaiting.

\rev{In order to show the effectiveness of our approach, we compare\footnote{The tests were run in a Python script for both methods.} the computation speed of C3 and the MIQP formulation (as in \eqref{eq:MPC_MIQP}) for different MPC horizons $N$. For the comparison experiments, we pick $s = 5, \rho = 1, \rho_s = 1.01$, and present the computation times (in seconds) in Table \ref{tab:finger_gaiting_horizon_comparison}. We also include the solution optimality comparison of those two methods where the optimality is compared by calculating the accumulated cost for a given time ($2$ seconds for this example).
The tabulated results show that C3 is substantially faster than a baseline MIQP solver, particularly as the horizon increases, and that this speed comes at only a modest increase in accumulated cost.}

\rev{
\begin{table}[t!]
	\centering
	\caption{Solve Time and Optimality Comparison for Finger Gaiting}
	\label{tab:finger_gaiting_horizon_comparison}
	\renewcommand{\arraystretch}{1.2}
    \begin{tabular}{|c|c|c|c|c|c|}
            \hline
            \multirow{2}{*}{\textbf{Item}}& \multirow{2}{*}{\textbf{Formulation}}  &\multicolumn{4}{c|}{\textbf{MPC Horizon} $N$} \\
			\cline{3-6}
            &		&		$10$&		$20$&		$30$&		$50$\\
			\hline
            \multirow{2}{*}{\textbf{Solve Time (s)}}&	\textbf{MIQP}& 		$0.112$&	$0.294$&	$0.791$&	$2.315$\\
			\cline{2-6}
			& 											\textbf{C3}&		$0.025$&	$0.051$&	$0.067$&	$0.095$\\
			\hline
            \multirow{2}{*}{\textbf{\shortstack{Accumulated \\Cost ($\times 10^{7}$)}}}&	\textbf{MIQP}& 		$3.467$&	$3.454$&	$3.451$&	$3.452$\\
			\cline{2-6}
			& 																				\textbf{C3}&		$4.979$&	$4.248$&	$4.101$&	$9.552$\\
			\hline
    \end{tabular}
\end{table}
}

\subsection{Pivoting}

\begin{figure}[t!]
	\hspace*{0.5cm}
	\includegraphics[width=0.8\columnwidth]{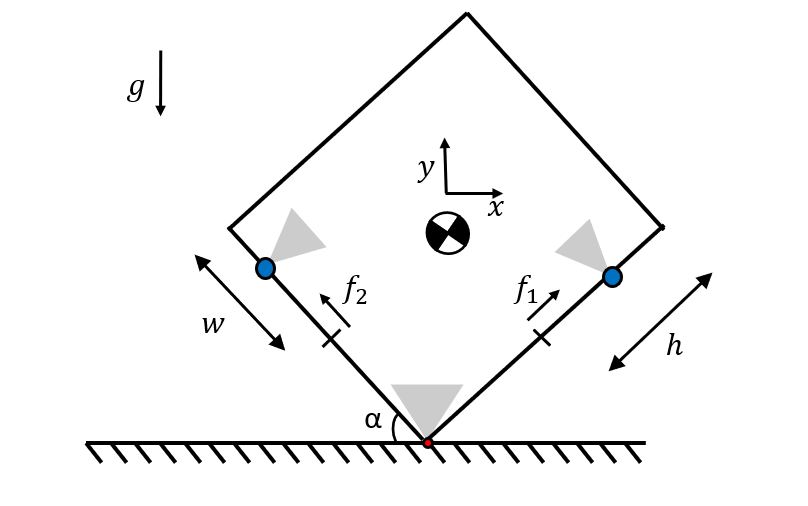}
	\caption{Pivoting a rigid object with two fingers (blue). The object can make and break contact with the ground and gray areas represent the friction cones.}
	\label{pivoting}
\end{figure}

Here, we want to demonstrate that our algorithm is capable of making decisions about stick-slip transitions as well as making/breaking contact. We consider pivoting a rigid object that can make and break contact with the ground inspired by Hogan et. al \cite{hogan2020tactile}. Two fingers (indicated via blue) interact with the object as in Figure \ref{pivoting}. The goal is to balance the rigid-object at the midpoint.

The positions of the fingers with respect to the object are described via $f_1$, $f_2$ respectively. The normal force that the fingers exert onto the box can be controlled. The center of mass position is denoted by $x$ and $y$ respectively, $\alpha$ denotes the angle with the ground and $w=1$, $h=1$ are the dimensions of the object. The coefficient of friction for the fingers are $\mu_1 = \mu_2 = 0.1$, and the coefficient of friction with the ground is $\mu_3 = 0.1$. We take the gravitational acceleration as $g=9.81$ and mass of the object as $m=1$. We model the system using an implicit time-stepping scheme \cite{stewart2000implicit}. \rev{The system has three contacts where the finger contacts are represented by $3$ complementarity variables each (1 slack variable and 2 frictional force variables), and the ground contact is modeled via $4$ complementarity variables (1 normal force variable, 1 slack variable and 2 frictional force variables). There are ten states ($n_x = 10$), including the pose and velocity of the cube and the fingers. Also, the system has ten complementarity variables ($n_\lambda = 10$), and 4 inputs ($n_u = 4$) that consist of the normal forces of the finger contacts and the acceleration of the fingers. We note that this system has $3\times 3\times 4 = 36$ hybrid modes.}

\begin{figure}[b!]
	\includegraphics[width=1\columnwidth]{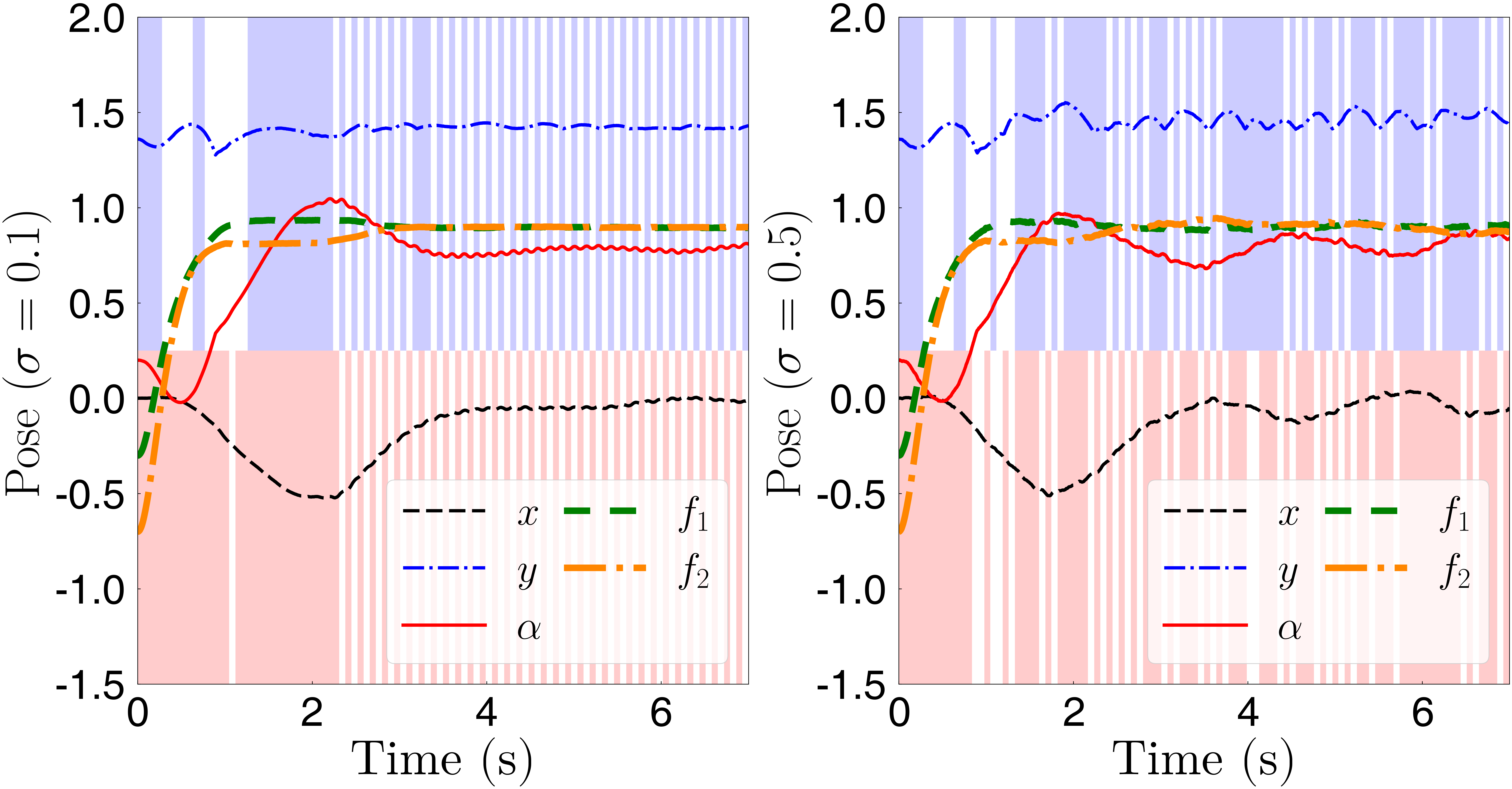}
	\caption{\rev{For the pivoting example, Gaussian process noise is added with with standard deviations $\sigma$. The upper blue shading implies that gripper 1 is applying normal force to the object whereas the lower red shading implies that gripper 2 is applying normal force.}}
	\label{pivoting_random}
\end{figure}

For this example, as the LCS-representation is only an approximation, we compute a new local LCS approximation at every time step $k$. We pick $s = 5$, $N = 10$, $\rho_s = 1.1$, \rev{$\Delta t = 0.01$ for $\mathcal{L}_{\Delta t}$} and use the local LCS approximation given at time-step $k$ while planning. \rev{With improvements, C3 runs at around $43.4$ Hz, an approximately 4x increase from the previously reported performance of 16 Hz \cite{aydinoglu2022real}.}

We present an example where the fingers start close to the pivot point where $f_1 = -0.3$, $f_2 = -0.7$ and the objects configuration is given by $x = 0$, $y=1.36$ and $\alpha = 0.2$. The goal is to balance the object at the midpoint ($x=0$, $y = \sqrt{2}$,  $\alpha = \pi/4$) while simultaneously moving the fingers towards the end of the object $(f_1 = f_2 = 0.9)$. Figure \ref{pivoting_random} demonstrates the robustness of the controller for different Gaussian disturbances (added to dynamics) with standard deviations (for $\sigma=0.1,0.5$). Note that at every time step, all positions and velocities (including angular) are affected by the process noise. The object approximately reaches the desired configuration (midpoint where $\alpha=\frac{\pi}{4}$) for $\sigma=0,0.05,0.1$ and starts failing to get close to the desired configuration for $\sigma=0.5$. Plots with $\sigma=0$ and $\sigma = 0.05$ are omitted as those were similar to the one with $\sigma = 0.1$. We emphasize that the process noise causes unplanned mode changes and the controller seamlessly reacts. \rev{We also observe that C3 succeeds across a wide range of frictional conditions; we sweep the ground contact parameter $\mu_3$ from $0.1$ to $1$ with $0.1$ increments, and observe that controller is always successful.} 
This example demonstrates that our method works well with successive linearizations as many multi-contact systems can not be captured via a single LCS approximation. \rev{Similar to the finger gaiting example, we report the speed and optimality comparison\footnote{The tests were run in a Python script for both methods.} between C3 and the MIQP formulation in Table \ref{tab:pivoting_horizon_comparison}. For this example, we calculate the accumulated cost until 1.5 $s$ with $s=5, \rho = 0.02, \rho_s = 1.1$. We can observe that with a horizon of $N=50$, it takes more than $300$ seconds for the MIQP formulation to compute the control input while C3 managed to solve the problem in less than $0.5$ seconds with only a minimal increase in accumulated cost.}

\rev{
\begin{table}[t!]
	\centering
	\caption{Solve Time and Optimality Comparison for Pivoting}
	\label{tab:pivoting_horizon_comparison}
	\renewcommand{\arraystretch}{1.2}
    \begin{tabular}{|c|c|c|c|c|c|}
            \hline
            \multirow{2}{*}{\textbf{Item}}& \multirow{2}{*}{\textbf{Formulation}}  &\multicolumn{4}{c|}{\textbf{MPC Horizon} $N$} \\
			\cline{3-6}
            &		&		$10$&		$20$&		$30$&		$50$\\
			\hline
            \multirow{2}{*}{\textbf{Solve Time (s)}}&	\textbf{MIQP}& 		$0.132$&	$0.376$&	$1.676$&	$>300$\\
			\cline{2-6}
			& 											\textbf{C3}&		$0.059$&	$0.111$&	$0.167$&	$0.334$\\
			\hline
            \multirow{2}{*}{\textbf{\shortstack{Accumulated \\Cost ($\times 10^{5}$)}}}&	\textbf{MIQP}& 		$1.214$&	$0.931$&	$0.865$&	-\\
			\cline{2-6}
			& 																				\textbf{C3}&		$1.217$&	$0.954$&	$0.951$&	$0.968$\\
			\hline
    \end{tabular}
\end{table}
}

\subsection{Cart-pole with Soft Walls}

\begin{table}[t!]
	\centering
	\caption{Projection Run-time and averaged cost-to-go}
	\label{tab:table_speed}
	\renewcommand{\arraystretch}{1.2}
		\begin{tabular}{|c|c|c|c|c|c|}
			\hline
			\textbf{Projection Method}       & \textbf{Mean} $\pm$ \textbf{Std} ($s$)  & \textbf{Cost} \\ \hline
			\textbf{LCP}   & $ 1.4 \cdot 10^{-5}$ $\pm$ $ 1.8 \cdot 10^{-6}$  & $22.09 $      \\ \hline
			\textbf{MIQP}   & $ 1 \cdot 10^{-3}$ $\pm$ $ 1.2 \cdot 10^{-4}$  & $24.98$     \\ \hline
			\textbf{ADMM}   & $ 5.3 \cdot 10^{-4}$ $\pm$ $3.3 \cdot 10^{-5}$   & $ 35.74 $     \\ \hline
		\end{tabular}
\end{table}

\begin{figure}[b!]
	\hspace*{1.3cm}
	\includegraphics[width=0.7\columnwidth]{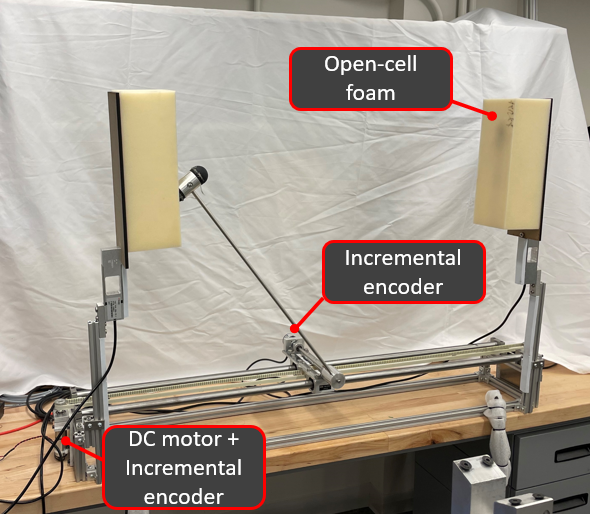}
	\caption{Experimental setup for cart-pole with soft walls.}
	\label{cartpole_experiment}
\end{figure}

\label{subsec:cartpole_sim}


In this subsection, we focus on an under-actuated, multi-contact system and demonstrate the effectiveness of our approach via hardware experiments. We consider a cart-pole that can interact with soft walls as in Figure \ref{cartpole_experiment}. This is a benchmark in contact-based control algorithms  \cite{marcucci2020warm, aydinoglu2020stabilization}. In the hardware setup, a DC motor with a belt drive generates the linear motion of the cart. Soft walls are made of open-cell polyurethane foam. The experiments in this subsection are done on a desktop computer with the processor Intel \emph{i7-6700HQ} and \emph{16GB RAM}.

First, numerical experiments are presented. Here, $x^{(1)}$ represents the position of the cart, $x^{(2)}$ represents the position of the pole and $x^{(3)}, x^{(4)}$ represent their velocities respectively. The forces that affect the pole are described by $\lambda^{(1)}$ and $\lambda^{(2)}$ for right and left walls respectively.

The model is linearized around $x^{(2)} = 0$ and $m_c=0.978$ is the mass of the cart, $m_p = 0.411$ is the combined mass of the pole and the rod, $l_p = 0.6$ is the length of the pole, $l_c = 0.4267$ is the length of the center of mass position, $k_1 = k_2 = 50$ are the stiffness parameter of the walls, $d=0.35$ is the distance between the origin and the soft walls. Dynamics are discretized using the explicit Euler method with time step $T_s = 0.01$ to obtain the system matrices and use the model in \cite{aydinoglu2020stabilization}. We note that the MIQP formulation (as in \eqref{eq:MPC_MIQP}) runs at approximately $10$ Hz.

We design a controller where $s = 10$, $\rho = 0.1$, $G_k = I$, $\rho_s = 2$, $N=10$, \rev{ and $\Delta t = 0.01$ for $\mathcal{L}_{\Delta t}$}. There is a clear trade-off between solve time and planning horizon \cite{li2021model}.
We test three projection methods described in Section \ref{sec:ADMM_main} and report run-times (single solve of \eqref{eq:approx_proj}) on Table \ref{tab:table_speed} averaged for $800$ solves.
We also report the average of cost-to-go value assuming all of the methods can run at $100$ Hz. Even though the MIQP projection is usually better at exploring new contacts, it does not always result in a better cost. We note that the controller can run slightly faster than $240$ Hz if the LCP-based projection is used. For this example, the QP in \eqref{eq:quadratic_program} has no inequality constraints and can therefore the KKT conditions can be directly solved.

Next, we test the multi-contact MPC algorithm on the experimental setup shown in Figure \ref{cartpole_experiment}. We consider the same LCS model with $k_1 = k_2 = 100$ as the stiffness parameters of the walls and $d = 0.39$ as the distance between the origin and walls.

\begin{figure}[t!]
	\includegraphics[width=1\columnwidth]{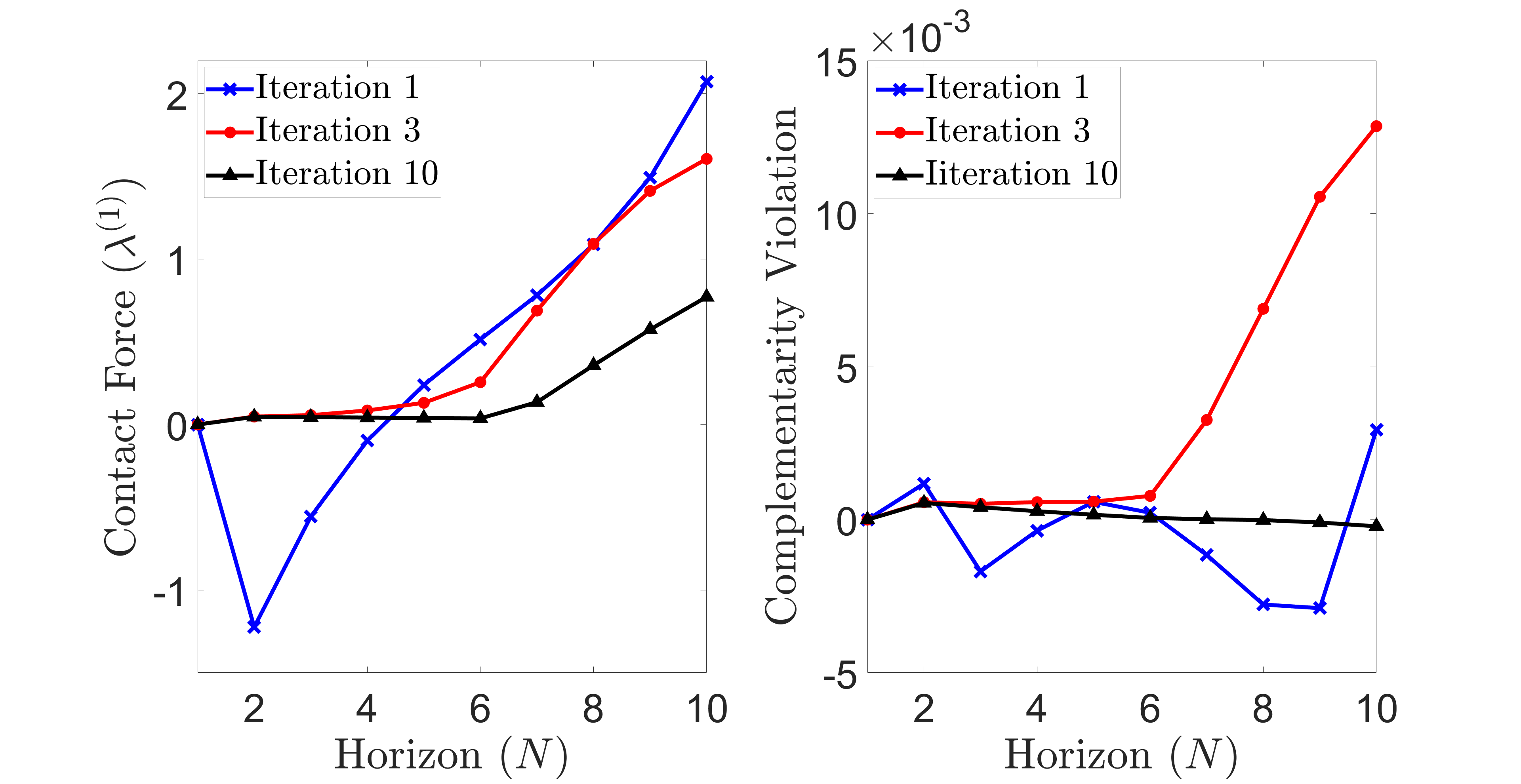}
	\caption{Evolution of contact force and complementarity violation during ADMM iteration when the cart is close to a contact surface. }
	\label{mm}
\end{figure}

\begin{figure}[b!]
	\includegraphics[width=1\columnwidth]{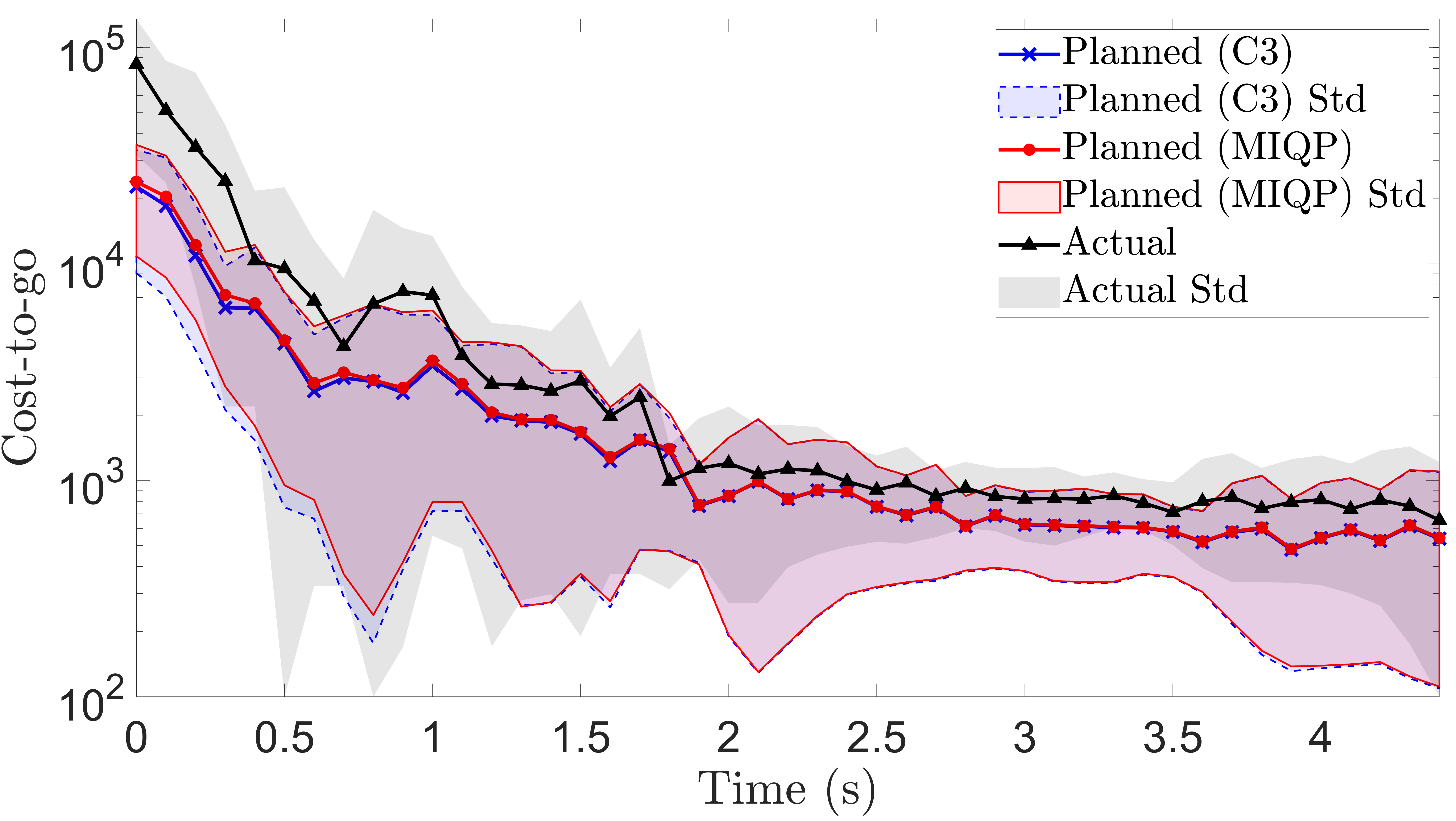}
	\caption{\rev{Approximate cost-to-go (cost as in \eqref{eq:MPC_MIQP}) values for the cart-pole experiment.}}
	\label{cartpole_cost_to_go}
\end{figure}

For the actual hardware experiments, the parameters of the MPC algorithm are $N = 10$, $\rho_s = 2.3$, $s = 10$, and $G_k = I$, $\rho = 0.5$. We use the LCP-based projection method, and solve the quadratic programs via utilizing the KKT system. While the algorithm is capable of running faster than $240$ Hz, due to the limited bandwidth between our motor controller and computer, we run the system at $100$ Hz. In Figure \ref{mm}, we illustrate, for one particular state, the evolution of the contact force throughout the ADMM process (at ADMM steps $1$, $3$, and $10$).
Notice that as the the algorithm progresses, complementarity violation decreases.

We initialize the cart at the origin and introduce random perturbations to cover a wide range of initial conditions that lead to contact events. Specifically, we start the cart-pole at the origin where our controller is active. Then, we apply an input disturbance $u_{\text{dist}} \sim U [10, 15]$ for $250$ ms to force contact events. We repeated this experiment $10$ times and our controller managed to stabilize the system in all trials. 

To empirically evaluate the gap between C3, which is sub-optimal, and true solutions to \eqref{eq:MPC_original}, and to assess the impact of modeling errors, we report the approximate cost-to-go values for our method, MIQP solution (as in \eqref{eq:MPC_MIQP}), and the actual observed states. 
More precisely, given the current state of the nonlinear plant, we calculate the cost as in \eqref{eq:MPC_original} using the inputs recovered from both the C3 algorithm and MIQP algorithm. Since C3 algorithm is not guaranteed to produce a $(x,u,\lambda)$ that strictly satisfies complementarity, the predicted cost may not match the simulated cost once $u$ is applied.
For the actual plant, we use the data from $N$ steps into the future and calculate the same cost. 
Notice that even though the cost-to-go of MIQP solution is always lower, as expected, the optimality gap is fairly small. 
This highlights that C3 finds near-optimal solutions, at least for this particular problem.
Also, notice that the actual cost-to-go matches the planned one which further motivates the applicability of LCS representations in model-based control for nonlinear multi-contact systems.

\section{Robot Arm Manipulation}

\label{sec:examples_high}

\begin{figure}[t!]
	\includegraphics[width=1\columnwidth]{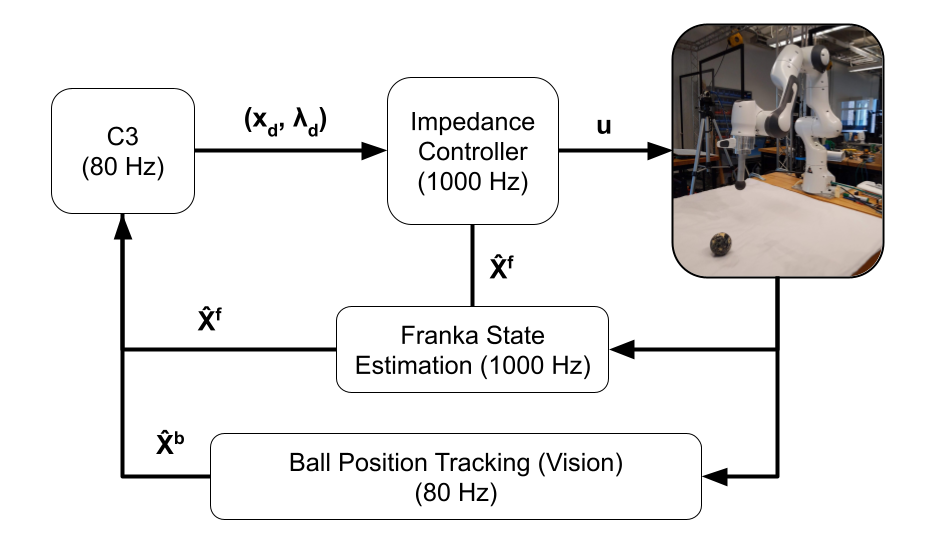}
	\caption{Key elements of the controller diagram.}
	\label{C3_high_level_fig}
\end{figure}




In this section, we demonstrate the effectiveness of our method on multiple robot arm manipulation tasks. In addition to that, we show that C3 can reliably be used as a high-level, real-time controller for multi-contact manipulation tasks that require high-speed reasoning about contact events. For these tasks, our goal is moving one or more rigid spheres along a desired trajectory using a Franka Emika Panda Arm with a spherical end-effector (Figure \ref{franka_ball_hw}). We note that the tasks involve multi-contact interaction between the arm and sphere(s). The robot has to decide on which side/location to make contact, and, in the multi-sphere case, which sphere to touch. 
Similar to the finger gaiting task in Section \ref{sec:examples_low}, and in contrast with comparatively simpler planar pushing tasks, these sphere-rolling problems require constantly making and breaking contact to reposition the hand. In addition to that the spheres are quite rigid, and they roll with very little dissipation. Hence it is critical for the controller to make rapid decisions to prevent them from escaping.

Towards this goal, we use the control architecture in Figure \ref{C3_high_level_fig}. \rev{In this scheme, C3 acts as a high-level controller and provides the desired state, contact force pair as proposed in Algorithm \ref{algortihm_high_level} (we note that one can choose to use simplified models if the model is high dimensional).} Then, a low-level impedance controller tracks these desired values. Please check the Appendix for details on Algorithm \ref{algortihm_high_level}, the simplified model and the impedance control scheme.

Both C3 and the impedance controller were implemented using the Drake toolbox \cite{drake}, and simulations were performed in Drake environment. OSQP \cite{osqp} is used to solve quadratic programs and Gurobi \cite{Gurobi} is used for mixed integer programs. The experiments are done on a desktop computer with the processor Intel \emph{i7-11800H} and \emph{16GB RAM}. Reported run-times include all steps in the algorithm. 


\begin{algorithm}[t!]
	\caption{}
	\begin{algorithmic}[1]
		\REQUIRE $\mathcal{M}, \theta$, $\Delta t$, $\hat{u}$
		 \STATE Estimate state as $\hat{x}$
		\STATE Obtain LCS model $\mathcal{L}_{\Delta t}$ around $(\hat{x}, \hat{u})$ using $\mathcal{M}$
		\STATE Run Algorithm \ref{algortihm_ADMM} with $\mathcal{L}_{\Delta t}$, $\theta$, $\hat{x}$ and obtain $u_{\text{C3}}$
		\STATE Compute $\Delta t_c $: Time spent during steps 1,2,3
		\RETURN $(x_d, \lambda_d)$ = $\mathcal{L}_{\Delta t_c} (\hat{x},  u_{\text{C3}} )$
	\end{algorithmic} 
	\label{algortihm_high_level}
\end{algorithm}

\subsection{Simulation example: Trajectory tracking with a single ball}
\label{sec:sim_experiment}

In this subsection, our goal is to move a rigid ball in a circular path inspired by works such as \cite{kurtz2022contact}. We also aim to make our simulation experiments as close to our hardware experiments (Section \ref{sec:state_based}) as possible. For the hardware experiments, the vision setup tracks position of the ball, but not orientation.
To emulate this, we measure the translational ball state $x^b_{x,y,z}$ at $80$ Hz rate and estimate translational velocity of the ball via finite differencing. Similarly, angular velocity is estimated by assuming that the ball is always rolling without slipping. The angular displacement is calculated by integrating the estimated angular velocity. 

To track a circular path, we form an LCS approximation around the current state and a quadratic cost for distance to a target state. The target (desired) state for the ball $x^b_d$ is calculated according to its current state  $x^b$.
Let $(x_c, y_c) = (0.55, 0)$ be the center of the desired ball path and $r=0.1$ be its radius. We define $\alpha^b$ as the current phase angle of the ball along its circular path, measured clockwise from the positive $y$-axis of Franka's base frame. More concretely, $\alpha^b = \mathrm{atan2}(x^b_x-x_c, x^b_y-y_c)$, where $x^b_x$ and $x^b_y$ are the current $x$ and $y$ coordinates of the ball respectively. Given the center of the path $(x_c, y_c)$, the radius of the path $r$, and the phase angle of the ball $\alpha^b$, we generate the next desired ball state $x^b_d$ as:


\begin{equation}
	\label{eq:state_based}
	\begin{aligned}
		& x^b_{d,x} =  x_c + r \sin (\alpha^b + \alpha^l), \\
		& x^b_{d,y} =  y_c + r \cos (\alpha^b + \alpha^l),
	\end{aligned}
\end{equation}
where $x^b_{d,x}$ is the desired $x$ coordinate of the ball, $x^b_{d,y}$ is the desired $y$ coordinate of the ball and $\alpha^l = 20^\mathrm{o}$ is a design parameter that controls how far ahead the desired ball state should be along the circle with respect to its current position.


\begin{figure}[t!]
	\hspace*{0.3cm}
	\includegraphics[width=0.9\columnwidth]{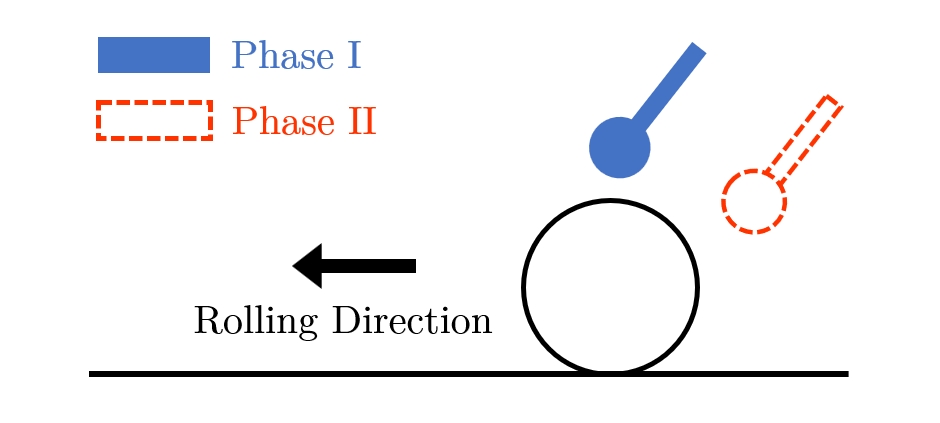}
	\caption{\rev{A cartoon illustrating the the time-based heuristic. The blue solid end-effector represents the end-effector bias in Phase I, and the red dashed end-effector represents the end-effector bias in Phase II.} }
	\label{heuristic}
\end{figure}

\begin{figure}[b!]
	\centering
	\includegraphics[width=1.\columnwidth]{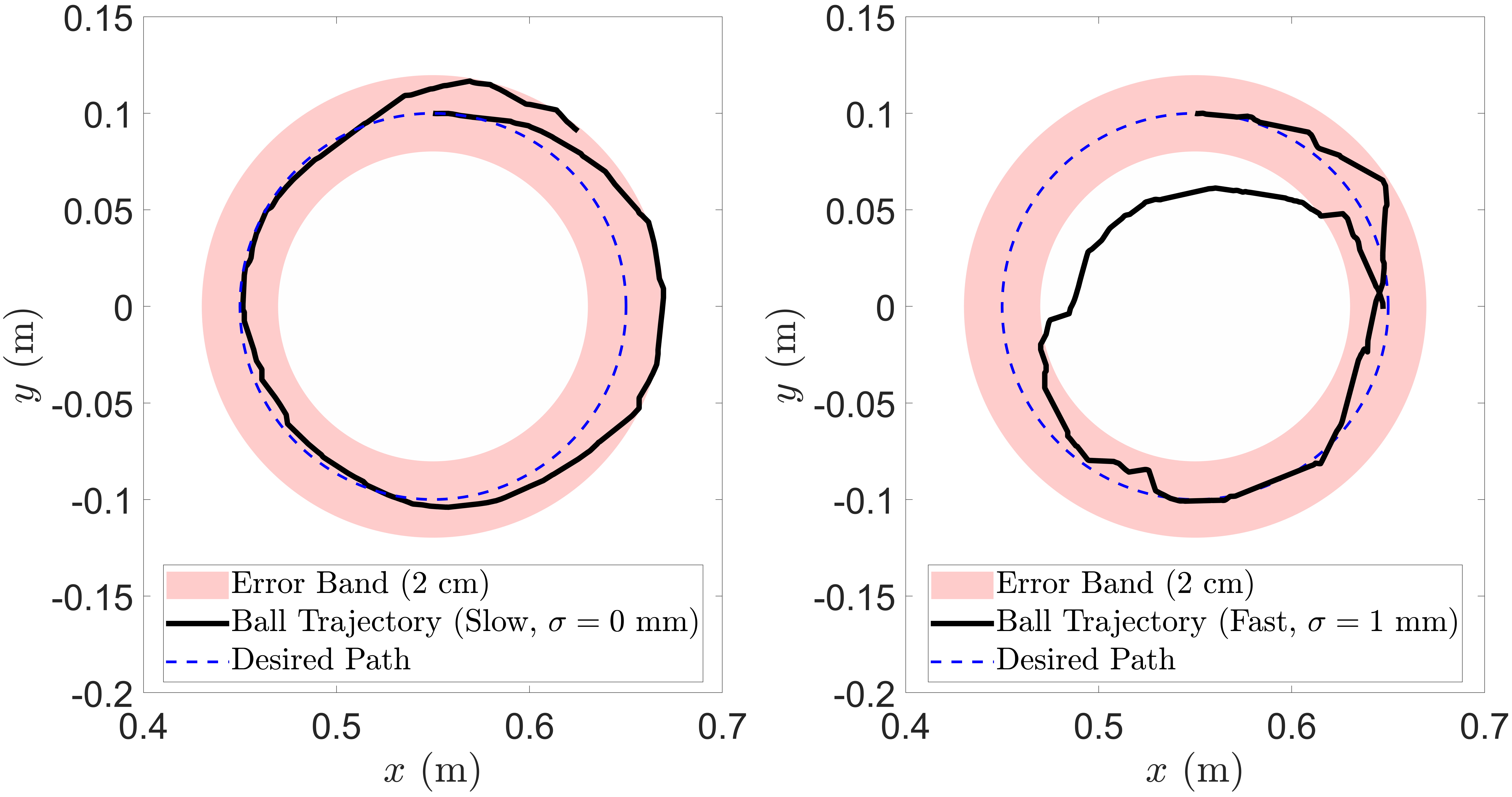}
	\caption{\rev{Trajectory tracking with single ball: Pink regions are the $2$ cm error bands, black lines represent the trajectory of the ball, and blue dashed lines represent the desired paths. Left shows the slow trajectory ($~\sim 30$ s for full circle) and right shows the fast trajectory ($~\sim 21$ s for full circle).}}
	\label{num_one_ball}
\end{figure}


One might easily warm-start C3 with an initial trajectory, and corresponding nominal hybrid plan. Here, to demonstrate that the algorithm can reliably discover trajectories in real-time from scratch, we explicitly do not provide such a warm-start.
Instead, we add a time-based heuristic as shown in Figure \ref{heuristic}. In Phase I ($1$ s long), we pick the end-effector bias slightly above the ball's current state (shown in blue). In Phase II ($0.5$ s long), we pick a set of end-effector biases so that it  ends up slightly behind the ball with respect to the desired ball location (shown in red) and also increase the cost on end-effector deviation. It is important to note that we do not specify when and how the contact interactions should happen. All the end-effector biases are above the ball, and algorithm decides that the end-effector should interact with the ball on its own (as there is also penalty on ball location error). Moreover, C3 can decide that an interaction should happen in either Phase I and Phase II, but we softly encourage interactions to happen in Phase I and the end-effector to reposition in Phase II.

The simplified model, $\mathcal{M}_s$, has nineteen states ($n_x = 19$), twelve complementarity variables ($n_\lambda = 12$) and three inputs ($n_u = 3$).
We design a high-level controller where $s = 2$, $\rho = 0.1$, $G_k = I$, $\rho_s = 3$, $\Delta t = 0.1$ and $N=5$. The controller, C3, can run around $73$ Hz \rev{(we note that solving the problem via MIQP approach as in \eqref{eq:MPC_MIQP} runs around $5$ Hz and fails to accomplish the task, also similarly fails when we early terminate the MIQP to make it faster)}. We strictly follow Algorithm \ref{algortihm_high_level} where a different LCS model $\mathcal{L}$ is generated using $\mathcal{M}_s$ to generate the desired state, contact force pair, $(x_d, \lambda_d)$. Afterwards, the impedance controller \eqref{eq:impedance} is used to track the the state, contact force pair as discussed.

We performed two different trials to demonstrate the accuracy of our approach as well its robustness against disturbances. For the first (Figure \ref{num_one_ball}, \rev{left}), we assume that there is no noise on ball position estimation. It is important to emphasize that there are still errors caused due to imperfect state estimation (as we emulate the hardware vision setup) but C3 is robust to these errors and the ball stays within the $2$ cm error band.

For the second example (Figure~\ref{num_one_ball}, \rev{right}), our goal is to demonstrate that C3 is robust and can recover even if things go wrong (e.g. ball goes out of $2$ cm error band). We add noise in state estimation of the ball (Gaussian noise with $1$ mm standard deviation) and increase the cost on desired ball location error to achieve faster motions.
Since we increase the cost on ball position error, C3 gives more aggressive commands to the low-level controller. Regardless, the ball stays in $2$ cm error band in roughly $75 \%$ of the motion. 
Even if the ball rolls away from the $2$ cm error band, the controller manages to recover and bring it back into the band utilizing its real-time planning capabilities. 
This further demonstrates the ability of C3 to recover from perturbations via re-planning contact sequences.
In the supplementary video, we present multiple circular trajectories tracked in a row (for both cases) to demonstrate the reliability of our approach.

\begin{figure}[t!]
	\centering
	\hspace*{0.3cm}
	\includegraphics[width=0.7\columnwidth]{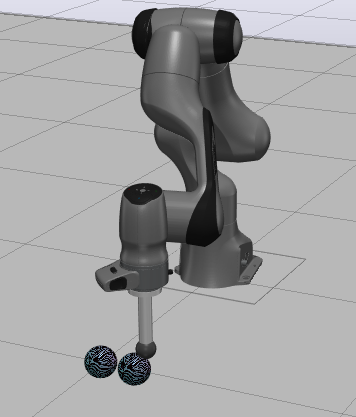}
	\caption{The Franka Emika Panda manipulating multiple rigid spheres.}
	\label{multiple_balls}
\end{figure}

\subsection{Simulation example: Trajectory tracking with multiple balls}

\label{subsec:multi_ball}

We consider a similar scenario to the previous example but with multiple balls (Figure \ref{multiple_balls}). The goal is to manipulate all of the balls to follow a line. Since the number of potential interactions (ball-ball and end-effector-ball interactions) are higher than the previous example, the hybrid decision making problem is significantly harder to solve. As the number of balls increase, the problem becomes high dimensional (with respect to both the state and the complementarity variable). To slightly reduce the dimension of this problem, we assume that the balls are always on the table and do not slide. We have tried this task both with $2$ balls and $3$ balls. The reduced-order model for $2$-ball setup has fourteen states ($n_x = 14$), eighteen complementarity variables ($n_\lambda = 18$) and three inputs ($n_u = 3$). For the 3-ball setup, the reduced-order model has eighteen states ($n_x = 18$), twenty four complementarity variables ($n_\lambda = 24$) and three inputs ($n_u = 3$).

For $i^\mathrm{th}$ ball, $x^{b_i}$, we pick the desired ball state as:
\begin{align*}
& x^{b_i}_{d,x} =  x_c, \\
& x^{b_i}_{d,y} =  x^{b_i}_{y} - d_y,
\end{align*}
where $x_c$ is a constant value and $d_y$ is the desired translation on $y$ direction. 
We also follow the same heuristic as the previous section and set the end-effector bias with respect to the ball that has made the least progress in the $y$ direction.


For the 2-ball setup, we design a controller that can run around $40$ Hz where $s = 2$, $\rho = 0.1$, $G_k = I$, $\rho_s = 3$, $\Delta t = 0.1$ and $N=5$. Similar to the previous section, we follow Algorithm \ref{algortihm_high_level} to generate desired states and forces ($x_d, \lambda_d$). Impedance controller \eqref{eq:impedance} is used to track the desired states, and feedforward torque term is taken as zero $(\tau_{ff} = 0)$ for this specific example.

\rev{We demonstrate the value of the convex projection (Section~\ref{subsec:projection}) on the 3-ball setup. Here, with identical parameters as in the 2-ball setting, the MIQP projection (with the Anitescu contact model from Section~\ref{subsec:Anitescu}) runs at approximately $30$ Hz. With the convex projection, the controller rate increases to $100$ Hz. While this projection is approximate, we note empirically that the resulting controller is high performance and succeeds at the task.}

It is important to emphasize that we do not warm-start our algorithm with an initial trajectory and C3 decides on contact interactions (e.g. which ball to interact with, the interaction between balls themselves) on its own. We have successfully accomplished the task with 2 balls where the controller runs around $40$ Hz. \rev{For the 3-ball setup, C3 with MIQP projection ran at $30$ Hz, which was empirically too slow to stably complete the task. By using the convex projection, we can increase the control rate to $100$ Hz, where the task was completed successfully.
A full video of both experiments are provided in the supplementary material.}

\subsection{Hardware experiment: State-based trajectory}
\label{sec:state_based}

In addition to providing numerical examples, we also performed the experiment from Section \ref{sec:sim_experiment} on hardware. As before, our goal is to roll the rigid ball in Figure \ref{franka_ball_hw} along a circular path of radius $0.1$ m, centered at $(x_c, y_c)$ in Franka's base frame. We use the same low-level impedance controller presented in \eqref{eq:impedance} and C3 as a high-level controller. The next desired state for the ball is computed using the same state-based method described in equation \eqref{eq:state_based}.

\begin{figure}[t!]
	\hspace*{0.3cm}
	\includegraphics[width=0.9\columnwidth]{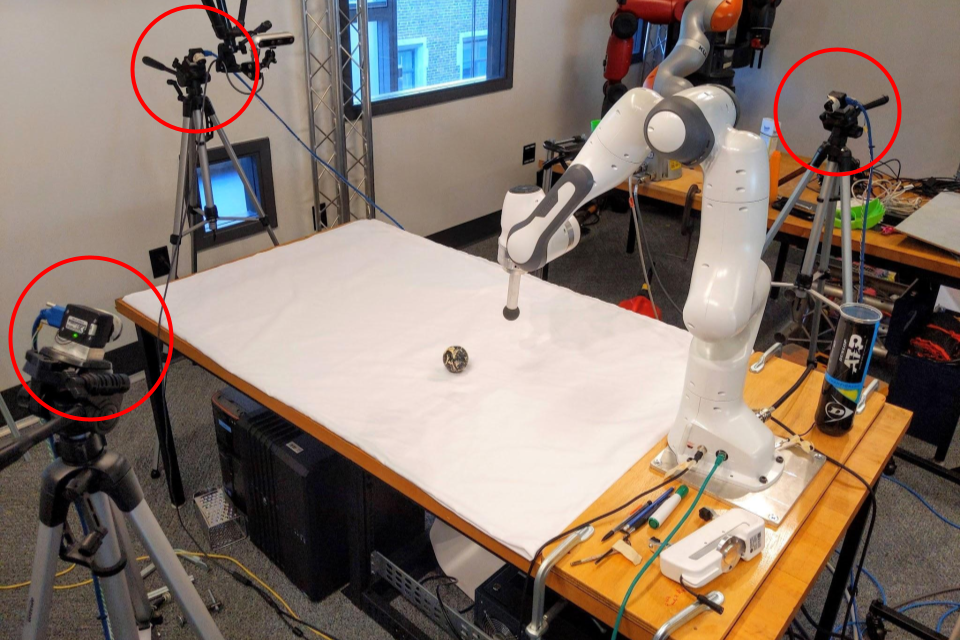}
	\caption{Camera setup for hardware experiments. The 3 PointGrey cameras are circled in red.}
	\label{fig:camera_setup}
\end{figure}

\begin{figure}[t!]
	\hspace*{-0.2cm}
	\includegraphics[width=1.05\columnwidth]{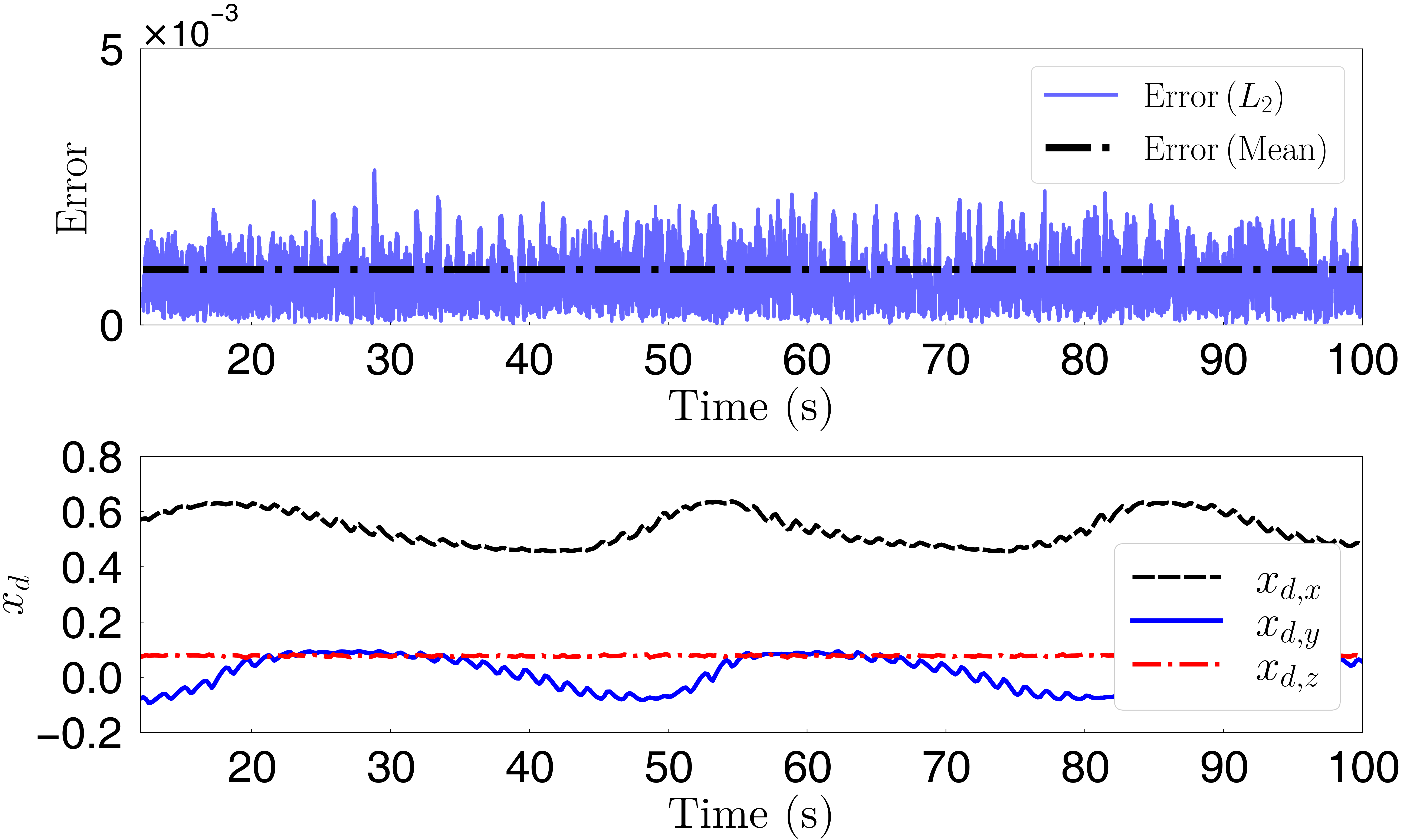}
	\caption{\rev{End-effector position tracking error ( $\sqrt{ \big ( \tilde{x}^e_x \big)^2 +  \big( \tilde{x}^e_y \big)^2 +  \big( \tilde{x}^e_z \big)^2  }$  ) and desired end-effector locations given by C3 ($x_d$) .}}
	\label{fig:error_desired}
\end{figure}

In order to complete this experiment on hardware, we need a vision pipeline to estimate the translational state of the ball, $x^b_{x,y,z}$. We accomplish this by positioning 3 PointGrey cameras around the robot as shown in Figure \ref{fig:camera_setup}. Each camera locates the ball using the Circle Hough Transform algorithm \cite{opencv_library},\cite{houghtransforms}, and computes an estimate of $x^b_{x,y,z}$ according to its extrinsic and intrinsic calibration. Next, the position estimates from all 3 cameras are compared, outliers are rejected, and the remaining estimates are averaged. The averaged measurements pass through a low-pass filter to produce the final estimate of the ball's translational state, $\hat{x}^b_{x,y,z}$. The ball's translational velocity, orientation, and angular velocity are estimated from $\hat{x}^b_{x,y,z}$ using the same procedure described in Section \ref{sec:sim_experiment}. The full vision pipeline operates at roughly 80-90 Hz and the position estimates typically vary between a range of $\pm 1$ mm.

The desired end-effector positions ($x^e_d$) that C3 generates as the high-level controller and the tracking error of the low-level impedance controller are shown in Figure \ref{fig:error_desired}. Note that the magnitude of the end-effector's translational error is consistently lower than $0.003$ m. 
These plots suggests that C3 produces effective trajectories that a low-level controller, such as our impedance controller, can realistically track.

\begin{figure}[b!]
	\hspace*{-0.5cm}
	\includegraphics[width=1\columnwidth]{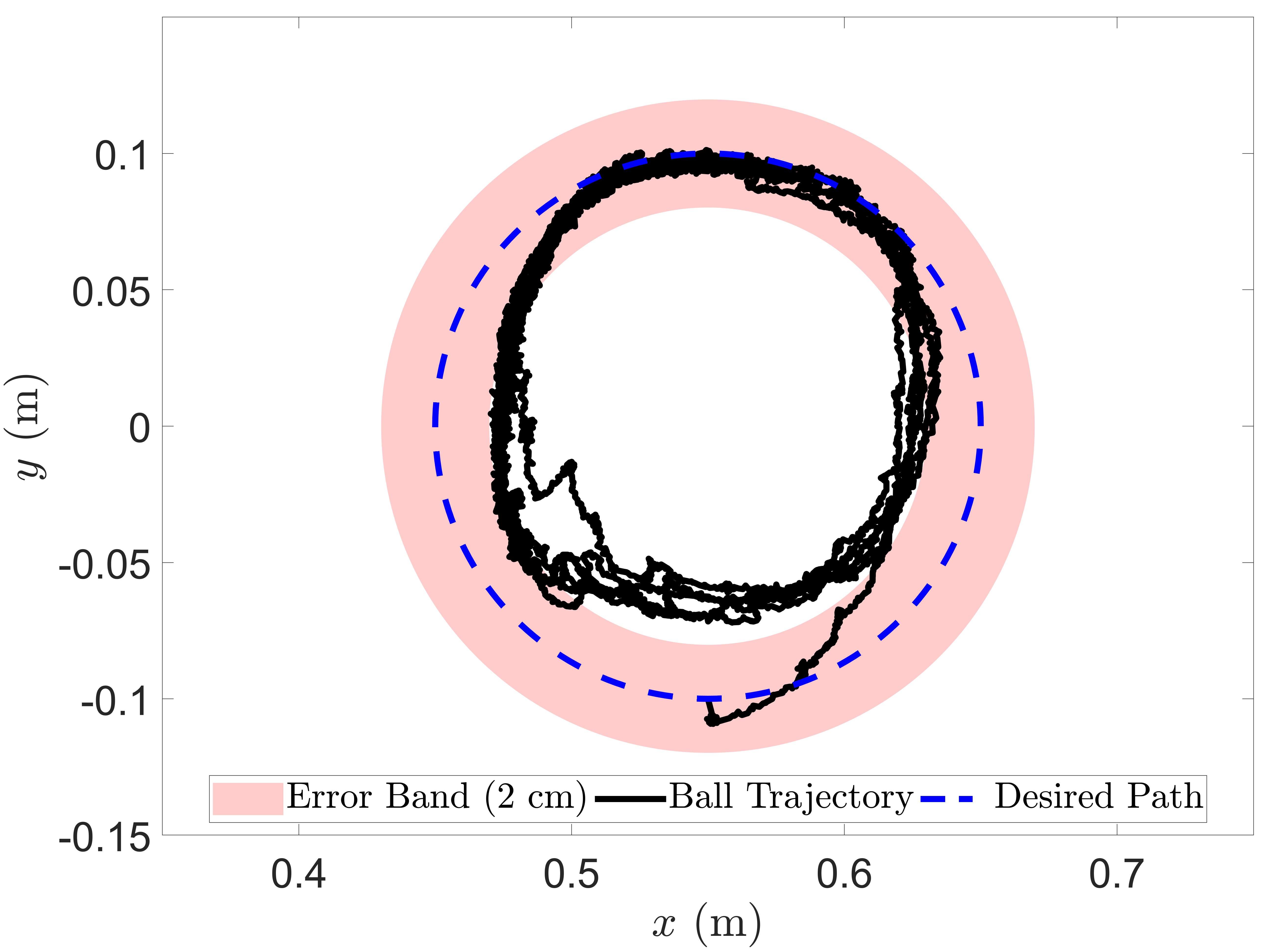}
	\caption{\rev{Hardware experiment: Tracking a state-based circular path. Pink region is the $2$ cm error band, black line demonstrates the actual trajectory of the ball, and blue dashed line represents the desired path.} }
	\label{fig:state_based_results}
\end{figure}

The trajectory of the ball is shown in Figure \ref{fig:state_based_results}, where the black line represents the ball path and the red line represents the desired trajectory. Each full rotation around the circle takes $35$ s on average to complete. We allow the robot to complete multiple loops to demonstrate the consistency of our approach. A full video of this experiment is provided in the supplementary material.

\begin{figure}[t!]
	\hspace*{0.3cm}
	\includegraphics[width=0.9\columnwidth]{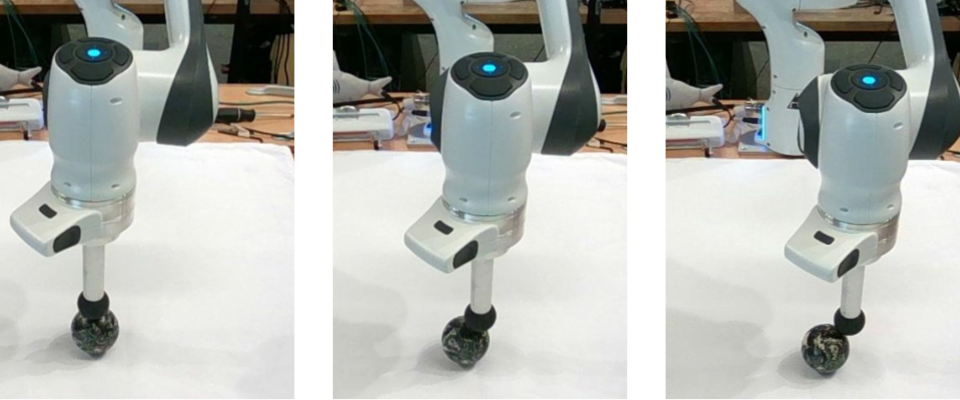}
	\caption{An example of a left to right roll using C3 as a high-level planner. Despite the provided heuristic, C3 positions the end-effector directly above the ball at the start of the motion (left image) and rolls from the front.}
	\label{fig:example_roll}
\end{figure}

\begin{figure*}[t!]
	\vspace*{-1cm}
	\includegraphics[width=2\columnwidth]{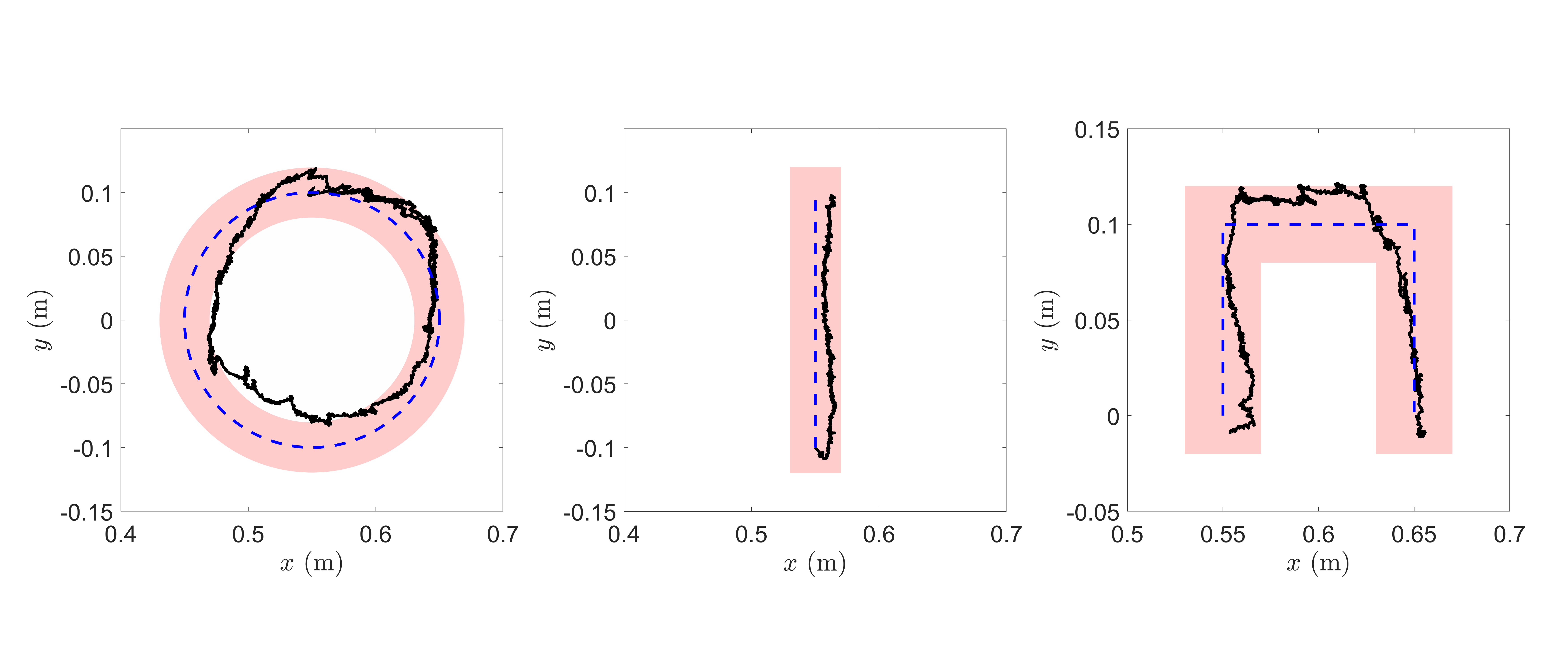}
	\vspace*{-1cm}
	\caption{\rev{Hardware experiment: Tracking time-based trajectories of different shapes. Pink regions are the $2$ cm error bands, black lines represent the states of the ball during the motion, and blue dashed lines represents the desired path.}}
	\label{fig:time-varying}
\end{figure*}

\begin{figure}[b!]
	\includegraphics[width=1\columnwidth]{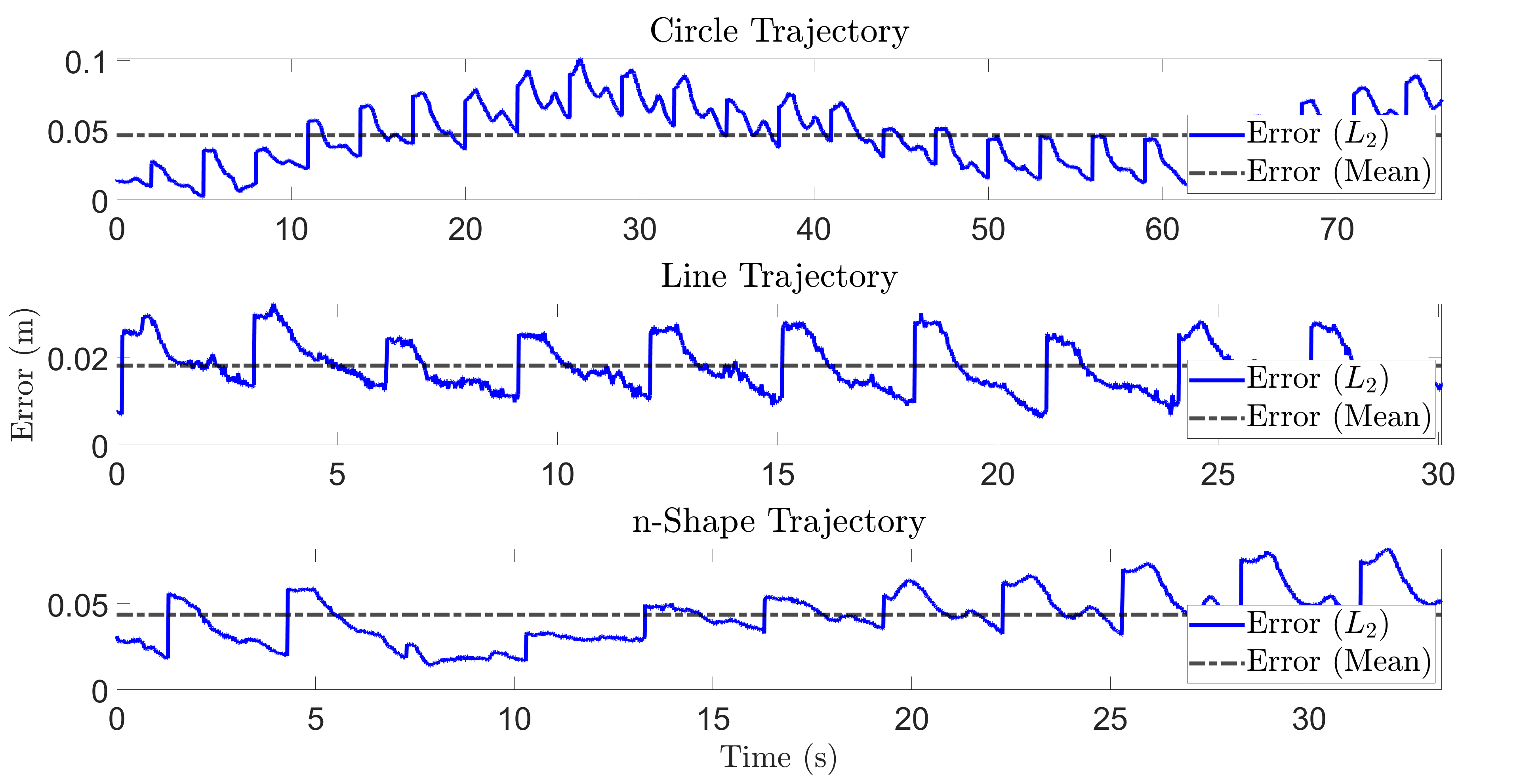}
	\caption{\rev{Ball position tracking error ($\sqrt{ \big ( \tilde{x}^b_x(t) \big)^2 +  \big( \tilde{x}^b_y(t) \big)^2}$ where $\tilde{x}^b(t) = {x}^b_d(t) - {x}^b(t)  $) for the three time based trajectories shown in Figure~\ref{fig:time-varying}.}}
	\label{fig:time_based error}
\end{figure}

In this experiment, we use the same time-based heuristic for end-effector biasing shown in Figure \ref{heuristic}. Although Phase II of this heuristic encourages the robot to roll the ball from the back, C3 can roll from the front instead (adapting to errors) during the actual experiments (see Figure \ref{fig:example_roll}). This re-emphasizes the fact that we do not specify when or how the robot should interact with the ball, or predefine any trajectory for the end-effector to follow. It also demonstrates that C3 does not require perfectly tailored a priori heuristics to perform well. This provides two major advantages. Firstly, it enables C3 to recover, even when its plans deviate from the provided heurisitc. Secondly, it eliminates the need for heavily-engineered heuristics, which may be difficult or time-consuming to develop in practise.

Since C3 plans in real-time ($70-80$ Hz), it also exhibits some robustness to model error. Recall that C3 plans using the simplified model, $\mathcal{M}_s$. This model assumes that the rolling surface is perfectly horizontal; however, the table in the hardware setup is slightly tilted.

This tilt is observable in the supplementary video: note that the ball tends to roll towards the right side of the camera frame whenever it is not in contact with the robot. As a result, the overall shape of the ball's trajectory in Figure \ref{fig:state_based_results} is circular, but it is shifted in the positive $y$ direction. Another potential source of model error in $\mathcal{M}_s$ is the friction coefficient between the ball and end-effector, which was simply set to $\mu=1$. Despite these model errors, C3 was still able to generate plans through contact to successfully and consistently complete the task.

Furthermore, the real-time planning capabilities of C3 enable it to continually correct any errors in the ball's trajectory. This may occur if the high-level controller generates an ineffective plan or if the low-level controller momentarily fails to track the end-effector trajectory. For instance, on one of the loops in Figure \ref{fig:state_based_results}, the robot erroneously rolls the ball to $(x^b_x, x^b_y) \approx (0.5, -0.02)$, nearly $5$ cm away from the desired path. Nonetheless, C3 is able to correct this mistake on the subsequent rolls and return the ball to its nominal trajectory. This demonstrates C3's self-correction abilities.

Lastly, we also demonstrate that C3 can account for external disturbances where we manually disturb the ball during the experiment. C3 accomplishes the task well even with such disturbances (shown in supplementary video).

\subsection{Hardware experiment: Time-based trajectory}

In this section, we conduct hardware experiments where the desired ball trajectory is parameterized as a function of time rather than the ball's current state. All other aspects of the experimental setup (the system models, high-level control algorithm, low-level controller, and vision pipeline) remain unchanged from Section \ref{sec:state_based}.

We experimented with the 3 different time-based trajectories, $x^b_d = x^b_d(t)$ shown in Figure \ref{fig:time-varying}, \rev{and the tracking error for these three trajectories are shown in Figure \ref{fig:time_based error}}. For all 3 time-based trajectories, C3 successfully and consistently discovers strategies to roll the ball around the $2$ cm error band. Additionally, the shape of the ball's trajectory closely matches the desired trajectory. These experiments demonstrate that C3 can also track time-based trajectories. Note that the same model errors (e.g. table slant, friction coefficient estimates, etc) from Section \ref{sec:state_based} are present in these experiments as well; however, since the high-level 
control algorithm has not changed, C3 remains sufficiently robust to successfully execute the tasks. The full videos of these experiments are available in the supplementary material.

\section{Conclusion}

In this work, we present an algorithm, C3, for model predictive control of multi-contact systems. The algorithm relies on solving QP's accompanied by projections, both of which can be solved efficiently for multi-contact systems. The effectiveness of our approach is verified on five numerical examples and our results are validated on two different experimental setups. We demonstrate that C3 can be reliably used both as a high-level and a low-level controller.
For fairly complex examples with frictional contact, our method has a fast run-time. We also demonstrate through experiments that our heuristic can find close-to-optimal strategies.

The framework tackles the hybrid MPC problem by shifting the complexity to the projection sub-problems. These projections can be difficult to solve, especially as we rely on the MIQP-based projection method for problems with frictional contact. Exploring alternate heuristics for the projection step is of future interest. Similarly, it would be interested to leverage learning-based approaches \cite{cauligi2020learning} to speed up the process.

As our approach is purely model-based, it is highly dependent on model parameters but some of those such as coefficient of friction can be hard to estimate accurately. Integration with learned models \cite{pfrommer2020contactnets}, hopefully in an adaptive way is in the scope of future work.

As discussed in Section \ref{sec:examples_high}, C3 generates desired contact forces but these have been used in a feedforward manner. It would be interesting to utilize tactile sensors and tactile feedback controllers \cite{aydinoglu2020stabilization, kim2023simultaneous} that track these desired contact forces. We believe that this can improve the robustness of our approach and is one of the crucial steps moving forward.

The choice of parameters (such as C3 parameters $\theta$) greatly affects the performance of the algorithm and exploring different approaches for deciding on parameters is of future interest too.


\section*{APPENDIX}
\subsection{ADMM}
We will clarify the derivation of  general augmented Lagrangian \eqref{eq:lagrangian} and ADMM iterations \eqref{eq:bir}, \eqref{eq:iki}, \eqref{eq:uc}. Consider the equivalent optimization problem to \eqref{eq:MPC_consensus}:
\begin{equation}
	\label{eq:MPC_consensus_appendix}
	\begin{aligned}
		\min_{ z } \quad & c(z) + \mathcal{I}_\mathcal{D} ( z  ) + \mathcal{I}_\mathcal{C} ( z  ) + \sum_{k=0}^{N-1} \mathcal{I}_{\mathcal{H}_k} (\delta_k) \\
		\textrm{s.t.} \quad &T (z - \delta) = 0
	\end{aligned}
\end{equation}
where $T = \mathbf{blkdiag}(T_0, \ldots, T_{N-1})$ and $T_k^T T_k = G_k$. The augmented Lagrangian is:
\begin{align*}
	\mathcal{L}_y & (z,  \delta, y) = c(z) + \mathcal{I}_\mathcal{D} ( z ) + \mathcal{I}_\mathcal{C} ( z )   \\
	&+ \sum_{k=0}^{N-1} \big( \mathcal{I}_{\mathcal{H}_k} (\delta_k) + y_k^T T_k (z_k - \delta_k) + \rho || T_k (z_k - \delta_k) ||_2^2  \big ),
\end{align*}
where $y^T = [y_0^T, y_1^T, \ldots, y_{N-1}^T]$, $y_k$ are the dual variables. In order to solve \eqref{eq:MPC_consensus_appendix}, ADMM iterations are:
\begin{align}
	& z^{i+1} = \text{argmin}_z \mathcal{L}_y (z, \delta^i, y^i), \\
	\label{eq:ikii}
	& \delta^{i+1} = \text{argmin}_{\delta}  \mathcal{L}_y (z^{i+1}, \delta, y^i), \\
	\label{eq:uci}
	& y^{i+1} = y^i + 2 \rho T (z^{i+1} - \delta^{i+1}).
\end{align}
Notice that individual vectors $\delta_k^{i+1}$ in \eqref{eq:ikii} can be computed as
\begin{equation*}
	\delta_k^{i+1} = \text{argmin}_{\delta_k}  \mathcal{L}_y^k (
	z^{i+1}_k, \delta_k, y_k^i)
\end{equation*}
due to separability of $\mathcal{L}_y$, where $\mathcal{L}_y^k(z_k, \delta_k, y_k) = \big( \mathcal{I}_{\mathcal{H}_k} (\delta_k) + y_k^T T_k (z_k - \delta_k) + \rho || T_k (z_k - \delta_k) ||_2^2  \big )$. Also note that $\delta_k^{i+1}$ only depends on $z_k^{i+1}$ and $y_k^i$. Similarly \eqref{eq:uci} can be written as
\begin{equation*}
	y^{i+1}_k = y^i_k + 2 \rho T_k (z^{i+1}_k - \delta^{i+1}_k).
\end{equation*}
After that, define the scaled dual variables $w_k$ such that $y_k = 2 \rho T_k w_k$. With $w_k$, the augmented Lagrangian is of the following form:
\begin{align*}
	\mathcal{L}_\rho & (z,  \delta, w) = c(z) + \mathcal{I}_\mathcal{D} ( z ) + \mathcal{I}_\mathcal{C} ( z )   \\
	&+ \sum_{k=0}^{N-1} \big( \mathcal{I}_{\mathcal{H}_k} (\delta_k) + 2 \rho w_k^T G_k (z_k - \delta_k) \\
	& + \rho (z_k - \delta_k)^T G_k (z_k - \delta_k)  \big )
	\end{align*}
following the fact that $G_k = T_k^T T_k$. It is equivalent to:
\begin{align*}
	\mathcal{L}_\rho & (z,  \delta, w) = c(z) + \mathcal{I}_\mathcal{D} ( z ) + \mathcal{I}_\mathcal{C} ( z )   \\
	&+ \sum_{k=0}^{N-1} \big( \mathcal{I}_{\mathcal{H}_k} (\delta_k) + \rho (r_k^T G_k r_k - w_k^T G_k w_k)  \big ).
\end{align*}
The corresponding ADMM iterations are:
\begin{align*}
	& z^{i+1} = \text{argmin}_z \mathcal{L}_\rho (z, \delta^i, w^i), \\
	& \delta_k^{i+1} = \text{argmin}_{\delta_k}  \mathcal{L}_\rho^k (z_k^{i+1}, \delta_k, w_k^i ), \; \forall k, \\
	& w_k^{i+1} = w_k^i + z^{i+1}_k - \delta_k^{i+1}, \; \forall k
\end{align*}
as $y_k = 2 \rho T_k w_k$ and $T_k$ is a positive definite matrix with $\mathcal{L}_\rho^k(z_k, \delta_k, w_k) =  \mathcal{I}_{\mathcal{H}_k} (\delta_k) + \rho (r_k^T G_k r_k - w_k^T G_k w_k)$.

\rev{
\subsection{Proof of Lemma \ref{lemma:convex_proj_err}}
The Lagrangian for the optimization problem \eqref{eq:convex_proj} is:
	\begin{align*}
		\mathcal{L} \left( \delta _{k}^{x},\delta _{k}^{\lambda}, \delta _{k}^{u}, \mu ,\eta \right) &=\frac{1}{2}\left\| \delta _{k}^{x}-x_d \right\| _{Q_x}+\frac{1}{2}\left\| \delta _{k}^{u}-u_d \right\| _{Q_u}
		\\
		&+\frac{\alpha}{2}\left\| \delta _{k}^{\lambda}-\lambda _d \right\| _F+\frac{1-\alpha}{2}\left\| \delta _{k}^{\lambda} \right\| _F
		\\
		&-\mu ^T\left( E\delta _{k}^{x}+F\delta _{k}^{\lambda}+H\delta _{k}^{u}+c \right) -\eta ^T\delta _{k}^{\lambda},
	\end{align*}
where $\mu \geq 0$ and $\eta \geq 0$. We consider the stationarity conditions:
\begin{align*} 
	&\frac{d\mathcal{L}}{d\delta _{k}^{x}}=0\implies Q_x\left( \delta _{k}^{x}-x_d \right) -E^T\mu =0,
	\\
	&\frac{d\mathcal{L}}{d\delta _{k}^{u}}=0\implies Q_u\left( \delta _{k}^{u}-u_d \right) -H^T\mu =0,
	\\
	&\frac{d\mathcal{L}}{d\delta _{k}^{\lambda}}=0\implies \alpha F\left( \delta _{k}^{\lambda}-\lambda _d \right) +\left( 1-\alpha \right) F\delta _{k}^{\lambda}-F\mu -\eta =0.
\end{align*}
It follows that:
\begin{align*} 
	&\delta _{k}^{x}=Q_{x}^{-1}E^T\mu +x_d,
	\\
	&\delta _{k}^{u}=Q_{u}^{-1}H^T\mu +u_d,
	\\
	&\delta _{k}^{\lambda}=\mu +F^{-1}\eta +\alpha \lambda _d.
\end{align*}
Primal and dual feasibility, and the complementary slackness constraints are:
\begin{align*}
	&E\delta _{k}^{x}+F\delta _{k}^{\lambda}+H\delta _{k}^{u}+c\ge 0,
	\\
	&\mu \ge 0,
	\\
	&\mu ^T\left( E\delta _{k}^{x}+F\delta _{k}^{\lambda}+H\delta _{k}^{u}+c \right) =0,
	\\
	&\eta \ge 0,
	\\
	&\delta _{k}^{\lambda}\ge 0,
	\\
	&\eta ^T\delta _{k}^{\lambda}=0.
\end{align*}
Let $\varTheta =E\delta _{k}^{x}+F\delta _{k}^{\lambda}+H\delta _{k}^{u}+c$, consider the complementarity error term from the original LCS \eqref{eq:LCS}:
\begin{equation}
	\label{eq:complementarity_error_bound}
	\begin{aligned}
		\operatorname{error_{comp}}&=(\delta _{k}^{\lambda})^T \varTheta =\left( \delta _{k}^{\lambda}-\mu +\mu \right) ^T\varTheta =\left( \delta _{k}^{\lambda}-\mu \right) ^T\varTheta 
		\\
		&\le \left\| \delta _{k}^{\lambda}-\mu \right\| \left\| \varTheta \right\|.
	\end{aligned}
\end{equation}
We wish to show that this error is governed by $\alpha$ (specifically, that it goes to 0 as $\alpha \to 0$). First, we analyze the term $\left\| \delta _{k}^{\lambda}-\mu \right\|$, and observe that:
\begin{equation*}
	\delta _{k}^{\lambda}-\mu =F^{-1}\eta +\alpha \lambda _d
	\implies\left\| \delta _{k}^{\lambda}-\mu \right\| \le \left\| F^{-1} \right\| \left\| \eta \right\| +\alpha \left\| \lambda _d \right\|.
\end{equation*}
From the complementarity slackness and stationarity, we have:
\begin{equation*}
	\eta ^T\delta _{k}^{\lambda}=\eta ^TF^{-1}\eta +\eta ^T\mu +\alpha \eta ^T\lambda _d=0,
\end{equation*}
and from dual feasibility we have:
\begin{equation*}
	\eta \ge 0,\mu \ge 0\implies \eta ^T\mu \ge 0.
\end{equation*}
Combining the two, it follows that:
\begin{equation*}
	\eta ^TF^{-1}\eta \le -\alpha \eta ^T\lambda _d.
\end{equation*}
Since $F$ is symmetric positive definite, it follows that:
\begin{equation*}
	\left\|\eta\right\| \leq \dfrac{\alpha \left\|\lambda_d\right\|}{\sigma_{\min}(F^{-1})},
\end{equation*}
and therefore that,
\begin{equation}
	\left\|F^{-1}\right\| \left\|\eta\right\| \leq \alpha \left\|\lambda_d\right\| \operatorname*{cond}(F^{-1}).
\end{equation}
Thus, we can conclude that
\begin{equation}
	\label{eq:lambda-mu_bound}
	\begin{aligned}
		\left\| \delta _{k}^{\lambda}-\mu \right\| &\le
		\alpha \left\|\lambda_d\right\| \left(1 + \operatorname*{cond}(F^{-1}) \right)
	\end{aligned}
\end{equation}
proving a linear bound in $\alpha$.
Next, we analyze the term $\left\| \varTheta \right\|$, expanding the term using the stationarity condition:
\begin{align*}
	\left\|\varTheta \right\| =\| (EQ_{x}^{-1}E\mu+&HQ_{u}^{-1}H+F)\mu +\eta
	\\
	&+Ex_d+\alpha F\lambda _d+Hu_d+c \|,
\end{align*}
so we need to analyze how $\mu$ behaves when $\alpha$ changes. 
While the primal solution is unique as the quadratic program is strictly convex, we do not assume LICQ and thus the optimal dual solution (e.g. $\mu$) may not be unique.
However, the optimal value can still be bounded. Since $F$ is positive definite, it follows that there exists a $\bar{\delta}_{k}^{\lambda}$ such that (\cite{cottle2009linear}, Lemma 3.1.3):
\begin{equation*}
	\bar{\delta}_{k}^{\lambda}>0, \quad F\bar{\delta}_{k}^{\lambda}>0.
\end{equation*}
Therefore, there exists some $\bar{\delta}_{k}^{x},\bar{\delta}_{k}^{\lambda}, \bar{\delta}_{k}^{u}$ where $\bar{\delta}_{k}^{\lambda}>0$ and $E\bar{\delta}_{k}^{x}+F\bar{\delta}_{k}^{\lambda}+H\bar{\delta}_{k}^{u}+c>0$.
With this, we prove the Slater condition and that strong duality holds. With a slight abuse of notation, let $p$ denote the primal objective function, $g$ denote the inequality constraints in the form $g \leq 0$, $d$ denote Lagrangian dual function with dual problem formulated as maximization, $M$ denote the optimal Lagrangian multiplier set. We have the following bound (\cite{Nedic2009bound}, Lemma 1):
\begin{equation*}
	\left\| \mu \right\| \le \max_{\left[ \mu ^T,\eta ^T \right] ^T\in M} \left\| \left[ \mu ^T,\eta ^T \right] ^T \right\| \le \frac{1}{\gamma}\left( p\left( \bar{\delta}_{k}^{x},\bar{\delta}_{k}^{u},\bar{\delta}_{k}^{\lambda} \right) -d\left( \mu ,\eta \right) \right),
\end{equation*}
where $\gamma =\min_{1\le i\le m} \left\{ -g_i\left( \bar{\delta}_{k}^{x},\bar{\delta}_{k}^{\lambda}, \bar{\delta}_{k}^{u} \right) \right\}>0$. In our context (optimization problem \eqref{eq:convex_proj}), we have:
\begin{align*}
	p\left( \bar{\delta}_{k}^{x},\bar{\delta}_{k}^{u},\bar{\delta}_{k}^{\lambda} \right) &=\frac{\alpha}{2}\left( \lambda _d-2\bar{\delta}_{k}^{x} \right) ^TF\lambda _d+\frac{1}{2}\left\| \bar{\delta}_{k}^{x}-x_d \right\| _{Q_x}
	\\
	&\qquad\qquad +\frac{1}{2}\left\| \bar{\delta}_{k}^{u}-u_d \right\| _{Q_u}+\frac{1}{2}\left\| \bar{\delta}_{k}^{\lambda} \right\| _F,
	\\
	g\left( \bar{\delta}_{k}^{x},\bar{\delta}_{k}^{\lambda},\bar{\delta}_{k}^{u} \right) &=\left[ -\left( E\bar{\delta}_{k}^{x}+F\bar{\delta}_{k}^{\lambda}+H\bar{\delta}_{k}^{u}+c \right) ^T, -\left( \bar{\delta}_{k}^{\lambda} \right) ^T \right] ^T.
\end{align*}
The point $(\bar{\delta}_{k}^{x},\bar{\delta}_{k}^{\lambda},\bar{\delta}_{k}^{u})$ is independent of $\alpha$ (which does not appear in the constraints), and can thus be regarded as constant for this analysis.
Since the primal objective is non-negative, $d(\mu, \eta) \geq 0$, and strong duality holds, the dual optimal value must be non-negative and we arrive at the bound
\begin{equation}
	\left\| \mu \right\| \le \frac{1}{\gamma}\left( \bar t_0 + \bar t_1\alpha \right) 
\end{equation}
for some $\bar t_0, \bar t_1\ge 0$.
And thus, that $\left\|\varTheta \right\|$ is similarly bounded
\begin{equation}
	\label{eq:Theta_bound}
	\begin{aligned}
		\left\|\varTheta \right\| &\leq t_0 + t_1\alpha,
	\end{aligned}
\end{equation}
for some $t_0, t_1\geq0$.
Combining (\ref{eq:complementarity_error_bound}), (\ref{eq:lambda-mu_bound}) and (\ref{eq:Theta_bound}), we have:
\begin{align}
	\operatorname{error_{comp}}&\le \alpha \left\|\lambda_d\right\| \left(1 + \operatorname*{cond}(F^{-1}) \right) \left(t_0 + t_1 \alpha \right).
\end{align}
Therefore, $\operatorname{error_{comp}}\rightarrow 0$ as $\alpha\rightarrow0$ with linear rate for small $\alpha$. $\qquad \qquad \qquad \qquad \qquad \qquad \qquad \qquad \qquad \qquad \qquad \quad \blacksquare$}
\subsection{Algorithm \ref{algortihm_high_level} and Simplified Model}

The algorithm requires a model $\mathcal{M}$ (Section \ref{sec:model}), C3 parameters $\theta$ (Section \ref{sec:C3}), discretization step length $\Delta t$, and a nominal input $\hat{u}$. First, an LCS approxmation ($\mathcal{L}_{\Delta t}$) of model $\mathcal{M}$ around the state estimate $\hat{x}$ and nominal input $\hat{u}$ is obtained. Then, (following Algorithm \ref{algortihm_ADMM}) one can use this LCS model along with the state estimate and pre-specified C3 parameters ($\theta$) and this process returns $u_{\text{C3}}$. Next, we compute the time spent ($\Delta t_c$) during the previous steps and calculate an LCS model $\mathcal{L}_{\Delta t_c}$ using $\Delta t_c$ as the discretization time-step. Lastly, we compute the desired state $x_d$ and the desired contact force $\lambda_d$ using $\mathcal{L}_{\Delta t_c}$, $\hat{x}$ and $u_{\text{C3}}$ (Section \ref{sec:model}).

Often times, C3 generates the desired end-effector trajectories and contact forces using a simplified model of the robot. Using such models for planning \cite{chen2020optimal} is extremely common and improves the solve time of the planning algorithms, e.g. C3. This specific simple model, $\mathcal{M}_s$, assumes one can control the translational accelerations of the spherical end-effector (Figure \ref{franka_ball_hw}) in all directions ($\ddot{x}^e_x,\ddot{x}^e_y,\ddot{x}^e_z$) and that the end-effector does not rotate.

\subsection{Impedance Control}

We use impedance control \cite{hogan1985impedance} to track the end-effector trajectories and contact forces that C3 produces which relies on model of Franka Emika Panda Arm, $\mathcal{M}_f$. Concretely, the controller we implemented is:
\begin{equation}
\label{eq:impedance}
\begin{aligned}
u = & J_f^T \Lambda \begin{bmatrix} K & 0 \\ 0 & B \end{bmatrix} \tilde{x}^e + C_f + \tau_{ff} \\ 
& + N_f \big( K_n (q_d^f - q^f) + B_n (\dot{q}_d^f - \dot{q}^f) \big)  ,
\end{aligned}
\end{equation}
where $x^e$ is the current state of the end-effector, $x_d^e$ is the desired end-effector state given by C3 and $\tilde{x}^e = x_d^e - x^e$ represents the error. The $K$ and $B$ are gain matrices for position error and velocity error respectively. Given $\mathcal{M}_f$, $J_f$ is the manipulator Jacobian for the end-effector and $\Lambda = \big( J_f M_f^{-1} J_f^T \big)^{-1}$ is the end-effector inertia matrix, where $M_f$ is the manipulator's mass matrix. The nullspace projection matrix $N_f$ directly follows from \cite{hermus2021exploiting}, $q_d^f$ and $\dot{q}_d^f$ are the desired null-space position and velocity for the manipulator, and $K_n$, $B_n$ are the corresponding gain matrices. The feedforward torque term is computed as $\tau_{ff} = J_c^T \lambda_d$ where $\lambda_d$ is the desired contact force obtained from C3 and $J_c$ is the contact Jacobian computed using the full-order model.

\section*{Acknowledgement}

We thank Philip Sieg and Terry Kientz for their help with the cart-pole experimental setup, Tianze Wang and Yike Li for their earlier explorations of ADMM-based multi-contact optimal control, Mathew Halm and William Yang for helpful discussions
related to the pivoting example and visualization code, Leon Kim, Bibit Bianchini and Peter Szczesniak for their help with the Franka experimental setup.

\bibliographystyle{ieeetr}
\bibliography{Refs}
\vspace{-12 mm}
\begin{IEEEbiography}[{\includegraphics[width=1in,height=1.25in,clip,keepaspectratio]{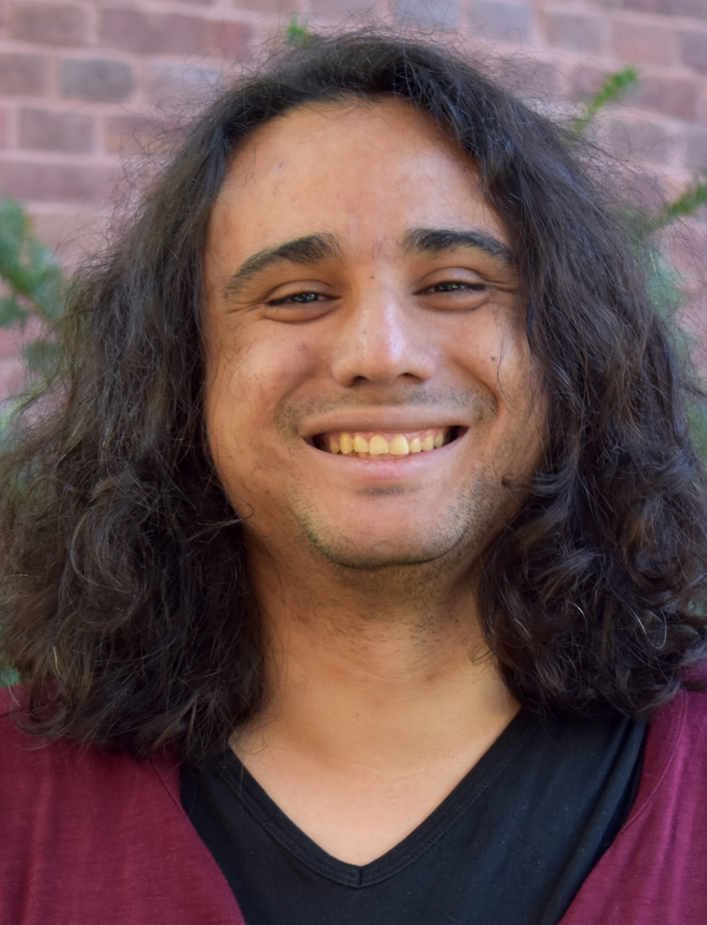}}]%
	{Alp Aydinoglu}
	completed his B.S. in Control Engineering from Istanbul Technical University in 2017 and is currently pursuing his Ph.D. in Electrical and Systems Engineering at University of Pennsylvania, working with Michael Posa in Dynamic Autonomy and Intelligent Robotics (DAIR) Lab. His research emphasizes control of multi-contact systems.
\end{IEEEbiography}
\vspace{-13 mm}
\begin{IEEEbiography}[{\includegraphics[width=1in,height=1.25in,clip,keepaspectratio]{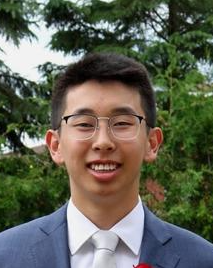}}]%
	{Adam Wei}
	received his B.ASc. in electrical engineering from the University of Toronto in 2019. He is currently pursuing a Ph.D. in robotics at the Massachussets Institute of Technology where he is advised by Prof. Russ Tedrake. His research interests include controls and imitation learning for robotic manipulation.
\end{IEEEbiography}

\begin{IEEEbiography}[{\includegraphics[width=1in,height=1.25in,clip, keepaspectratio]{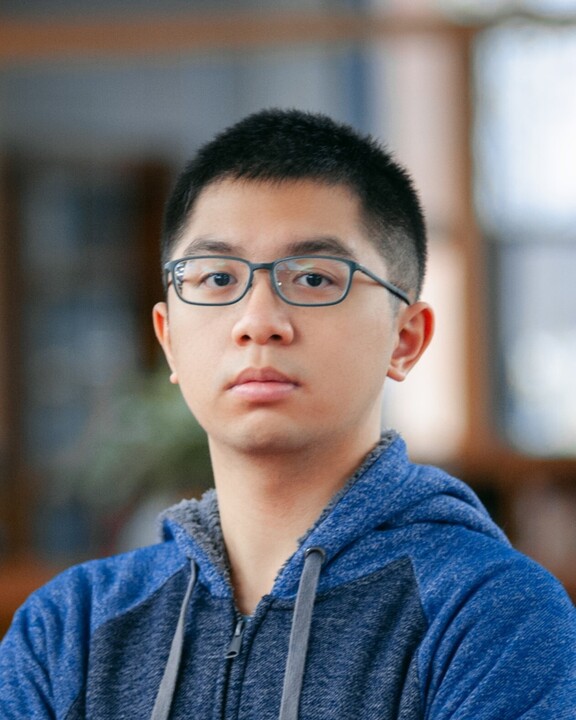}}]%
	{Wei-Cheng Huang}
	received his B.E. degree in Mechanical Engineering from Shanghai Jiao Tong University, Shanghai, China, in 2022, and his M.S.E. degree in Robotics from the General Robotics, Automation, Sensing and Perception (GRASP) Lab, University of Pennsylvania, Philadelphia, PA, USA, in 2024. He is currently pursuing his Ph.D. degree in Computer Science, advised by Professor Kris Hauser, at the University of Illinois Urbana-Champaign, Urbana, IL, USA.
	His research interests primarily focus on planning and control for contact-rich dexterous robotic manipulation. 
\end{IEEEbiography}
\vspace{-150 mm}
\begin{IEEEbiography}[{\includegraphics[width=1in,height=1.25in,clip, keepaspectratio]{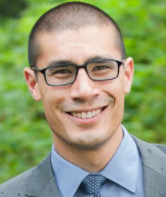}}]%
	{Michael Posa}
	received the B.S. and M.S. degrees in mechanical engineering from Stanford University, Stanford, CA, USA, in 2007 and 2008, respectively, and the Ph.D. degree in electrical engineering and computer science from the Massachusetts Institute of Technology, Cambridge, MA, USA, in 2017.,He is currently an Assistant Professor of mechanical engineering and applied mechanics with the University of Pennsylvania, Philadelphia, PA, USA, where he is a Member of the General Robotics, Automation, Sensing and Perception (GRASP) Lab. He holds secondary appointments in electrical and systems engineering and in computer and information science. He leads the Dynamic Autonomy and Intelligent Robotics Lab, University of Pennsylvania, which focuses on developing computationally tractable algorithms to enable robots to operate both dynamically and safely as they maneuver through and interact with their environments, with applications including legged locomotion and manipulation.,Dr. Posa was the recipient of multiple awards, including the NSF CAREER Award, RSS Early Career Spotlight Award, and Best Paper Awards. He is an Associate Editor for IEEE Transactions on Robotics.
\end{IEEEbiography}

\end{document}